%% file: main.tex
\documentclass{article}

\PassOptionsToPackage{sort}{natbib}

\usepackage[final]{corl_2026} 
\usepackage{hyperref}       
\usepackage{url}            
\usepackage{booktabs}       
\usepackage{amsfonts}       
\usepackage{nicefrac}       
\usepackage{microtype}      

\usepackage{amsmath}
\usepackage{enumitem}
\usepackage{multirow}
\usepackage{multicol}

\usepackage{colortbl}
\usepackage{adjustbox}
\usepackage{algorithm}
\usepackage{algorithmic}
\usepackage{subfig}
\usepackage{wrapfig}
\usepackage[separate-uncertainty=true]{siunitx}
\newcommand{\METHOD}{\underline{F}ailure-\underline{A}ware \underline{R}etry}
\newcommand{\method}{FAR}

\definecolor{lightgray}{rgb}{0.95, 0.95, 0.95}

\title{FAR: Failure-Aware Retry for Test-Time Recovery and Continual Policy Improvement}

\hypersetup{
  pdftitle={FAR: Failure-Aware Retry for Test-Time Recovery and Continual Policy Improvement},
  pdfauthor={Haoran Hao, Shahram Najam Syed, Jeffrey Ichnowski, Jeff Schneider}
}

\author{
  Haoran Hao
  \quad
  Shahram Najam Syed
  \quad
  Jeffrey Ichnowski
  \quad
  Jeff Schneider \\ 
  Robotics Institute, Carnegie Mellon University \\
}

\begin{document}
\maketitle


\begin{abstract}
Robot policies inevitably encounter failures when deployed in real environments.
Naive retries often repeat the same mistakes, while many existing recovery methods rely on human intervention. In this paper, we propose {\METHOD} ({\method}), a framework that enables robots to learn from previous failures at test time, adapt their behavior accordingly, and eventually complete the task autonomously.
\method{} combines Failure-Contrastive Preference Adaptation, which constructs preference learning data from failures to steer the policy away from previously unsuccessful behaviors, with lightweight action perturbations during retries to encourage local exploration.
We further incorporate successful recovery trajectories into a training loop for continual policy improvement. Experiments in both simulation and real-world manipulation tasks show that \method{} substantially improves success rates and robustness, with average gains of 17.6\% over the standard diffusion policy in simulation and 11.7\% in the real world. In addition, \method{} improves data efficiency under both reset and timestep budgets during continual policy improvement by exploiting informative failure cases.
Videos and code are available at \href{https://hoar012.github.io/FAR-Project/}{hoar012.github.io/FAR-Project}.
\end{abstract}



\section{Introduction}
Robot policies trained on offline expert demonstrations have achieved strong performance on a wide range of manipulation tasks \citep{chi2023diffusionpolicy, kim24openvla, black2025pi_05, Zhao-RSS-23-aloha}. However, when deployed in real environments, these policies inevitably encounter failures, such as missing the target during grasping or dropping an object during execution. Recovering from such failures is challenging because these states are rarely covered by offline demonstrations and are often out-of-distribution for the pretrained policy. As a result, simply re-executing the same policy often repeats the same mistake.

\begin{figure*}[t]
  \begin{center}
    \centerline{\includegraphics[width=\linewidth]{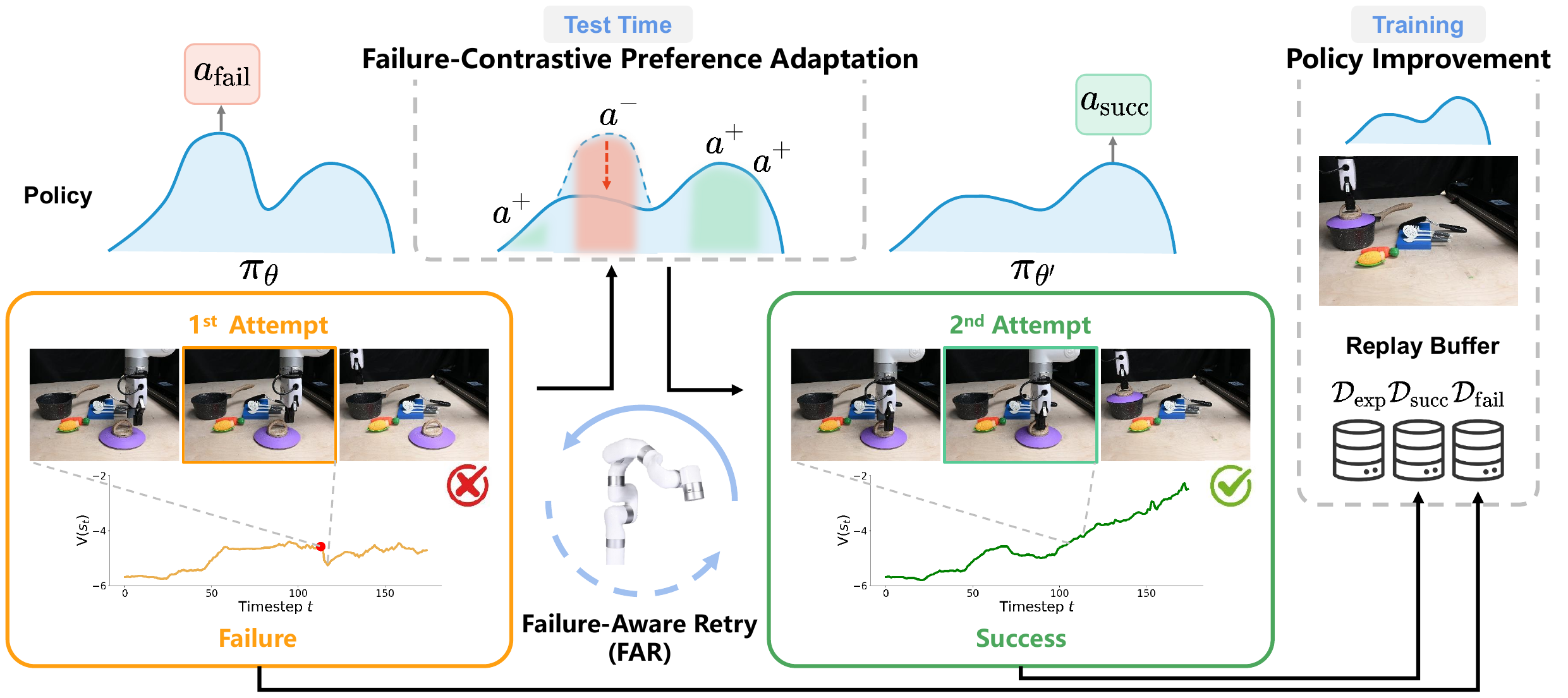}}
    \caption{
    \textbf{Overall Framework of {\method}.}
    After a failure, {\method} identifies failure-inducing actions using value estimation, then updates the policy with both failure examples and alternative positive examples. The collected trajectories are added to the replay buffer for continual policy improvement.
    }
    \vspace{-8mm}
    \label{fig:framework}
  \end{center}
\end{figure*}

Existing methods often rely on human intervention to correct robot behavior and collect additional data for future training \citep{intelligence2025pi06vla, hu2025RaC, luo2025precise}, including DAgger~\citep{pmlr-v15-ross11a-dagger} and related human-in-the-loop methods \citep{mandlekar2023humanintheloop, hoque2023fleet, hoque2021lazydagger, hoque2021thriftydagger}. 
While effective, they require substantial human effort. 
This raises the question of whether a robot can learn from failures, adapt its behavior at test time, and recover autonomously.

In this work, we propose a framework for learning from failure and adapting robot behavior at test time to improve retry-based recovery \citep{ebert2018robustness, du2024err}. Effective retries face two key challenges: the policy tends to repeat the same mistake after failure, and its action distribution comes from offline demonstrations, limiting exploration in out-of-distribution recovery states. Our approach addresses these challenges with Failure-Contrastive Preference Adaptation (FCPA), which constructs preference learning signals from failure experience to steer the policy away from previously unsuccessful behaviors, and action perturbations during retries to encourage local exploration. Together, they improve retry success and generate diverse recovery trajectories for continual policy improvement.

These recovery trajectories are valuable because they reveal the limitations and failure boundaries of the current policy, providing supervision that is absent from offline training. We therefore use them to further improve the policy. 
Starting from a policy pretrained on offline demonstrations, the robot
interacts with the environment and makes multiple attempts on each task using {\method}. 
The resulting trajectories are incorporated into online finetuning, allowing the policy to learn from challenging failure cases, improve recovery performance, and achieve greater data efficiency.

We evaluate {\method} on three simulation benchmarks spanning nine robot manipulation tasks,
as well as three real-world tasks on an xArm platform.
Our contributions are as follows:
(1) We introduce {\METHOD} ({\method}), a retry-based test-time recovery framework that combines failure-aware adaptation with lightweight exploration for more effective failure recovery.
(2) We show that {\method} generates informative recovery trajectories that provide useful supervision for continual policy improvement.
(3) We demonstrate that {\method} improves task success rates in both simulation and the real world, while reducing the need for costly environment resets during online learning.

\section{Related Work}
\textbf{Robot Failure Recovery.}
Robots need to handle a range of failures in order to operate robustly after deployment, since failures are inevitable in real-world environments \citep{xu2026agentchord, liu2023reflect}. These failures often arise from the limited fidelity of the learned policy or insufficient training data. 
For example, the robot may knock over a can and then fail to pick it up.
Several works \citep{gu2026safe, duan2025aha, ifailsense2026} use pretrained vision-language models (VLMs) to detect failures. Other methods use natural language reasoning to analyze failures \citep{duan2025aha} and provide corrections \citep{liu2023reflect, roboFAC, lin2025failsafe, vlm_recovery, hong2026learning}. 
There are also studies \citep{recovery_RL, xiao2026selfimproving, li2026failureawarerl, gupta2021reset} that use reinforcement learning to train policies to recover from unsafe or out-of-distribution states.
More recently, inspired by DAgger~\citep{pmlr-v15-ross11a-dagger}, several works \citep{intelligence2025pi06vla, hu2025RaC, luo2025precise, xu2025compliant, hoque2021lazydagger, hoque2021thriftydagger, mandlekar2023humanintheloop, hoque2023fleet} have incorporated human feedback to correct the robot and teach it how to recover from failures.
Our setting is closely related to prior retry-based recovery methods \citep{ebert2018robustness, du2024err}, where the robot is allowed multiple attempts after a failure. In particular, Bellman-Guided Retrials \citep{du2024err} detects discrepancies between expected and observed task progress and skews action sampling during retry. Unlike existing non-parametric retry strategies, we study test-time learning from failure experience, enabling the policy to improve subsequent retries using immediate interaction feedback without human intervention.

\textbf{Test-Time Adaptation.}
Test-time adaptation (TTA) aims to adapt a pretrained model to unlabeled test data using information available during inference \citep{liang2025comprehensive, wang2025search}. It has been widely studied in machine learning and language modeling, especially for domain adaptation \citep{chen2022contrastive, pmlr-TTT, yoon2024ctpt, NEURIPS2021_test_time_classifier, chen2026testtime, hu2025testtime, niu2024test}.
Recent work has also explored TTA in robotics \citep{kim2025testtime, wagenmaker2025behavioral, yoo2025world, liulocoformer}. For instance, \citet{liulocoformer} adapt locomotion models online to different robot embodiments and dynamics. \citet{bai2025evolvevla} use a progress estimator to provide dense feedback for adapting pretrained vision-language-action (VLA) models. \citet{wagenmaker2025behavioral} study in-context adaptation for more controllable exploration.
In this work, we focus on adapting a policy from previous failed attempts without resetting the task, enabling failure-aware retry and recovery.

\textbf{Offline-to-Online RL.}
Offline-to-online RL improves policies pretrained on offline data through additional environment interaction \citep{xie2021policy, zhang2023policy, lee2022offline, ball2023efficient, nakamoto2023calql, li2023accelerating, li2026reinforcement}. Prior work shows that continuing training with both offline and newly collected data is effective across different policy classes, including diffusion and flow-based policies \citep{dppo2024, park2025flow, zhang2026reinflow}, and has recently been extended to improve pretrained VLA models \citep{liu2026what, li2026simplevlarl, xiao2026selfimproving, vla_online_rl}. 
However, online improvement is often data- and labor-intensive, requiring many interactions and frequent environment resets \cite{gupta2021reset, mahmood2018benchmarking}. Standard data collection typically resets the environment once the policy fails within a fixed horizon, which exposes the model to diverse initial states but underutilizes hard failure cases that reveal the boundary of the current policy. 
Our work instead focuses on learning from these failure states, enabling the policy to improve from challenging situations while reducing costly resets and human effort.

\section{Problem Statement}
\textbf{Markov Decision Process.} We consider a Markov decision process (MDP)
$
\mathcal{M} = (\mathcal{S}, \mathcal{A}, P, p_0, r, \gamma),
$
where $\mathcal{S}$ and $\mathcal{A}$ are the state and action spaces, $P$ is the transition function, $p_0$ is the initial-state distribution, $r$ is the reward function, and $\gamma$ is the discount factor. A policy $\pi_\theta(a \mid s)$ maps states to actions, with the goal of maximizing expected discounted return.

\textbf{Generative Policies.}
We instantiate the policy in our framework using Diffusion Policy \citep{chi2023diffusionpolicy}. As a generative policy, it models a conditional distribution over action sequences, $\pi_\theta(a_{1:H}\mid s)$, enabling multi-modal behaviors. 
During training, Gaussian noise is added to expert action sequences, and the model learns to predict the injected noise conditioned on the state $s$. 
At inference time, the policy starts from Gaussian noise and iteratively denoises it to generate an action sequence.

\textbf{Setting.}
We study failure recovery for a robot policy pretrained on offline expert demonstrations. At deployment, after a failure, we preserve the scene state, return the robot to a predefined start pose, and allow multiple retries instead of resetting the environment. 
The objective is to adapt at test time from recent failures and improve the probability of success in subsequent retries.
Reward signals are used for critic training. During test-time adaptation, however, \method{} does not require access to any task-specific dense reward function. Instead, it relies only on task-level outcome feedback, such as whether an attempt succeeds or fails, together with recent interaction history.

\section{Method}
Effective retry requires the policy to both learn from previous failures and explore alternative actions in out-of-distribution states. To improve retry performance at test time, {\method} consists of two components: (1) Failure-Contrastive Preference Adaptation (FCPA), which updates the policy using failure feedback to improve subsequent attempts, and (2) lightweight action perturbation, which encourages local exploration to expand the policy support for harder recovery cases. Section~\ref{sec:far} presents these two components, and Section~\ref{sec:online} shows how the resulting recovery trajectories can be incorporated for continual policy improvement. The overall framework is shown in Figure~\ref{fig:framework}. 

\subsection{Failure-Aware Retry}
\label{sec:far}
\textbf{Conservative Value Estimation.}
{\method} uses value estimates to identify failure-inducing actions and construct data for adaptation. 
A key challenge is that failure states encountered at deployment are often out-of-distribution with respect to the offline demonstrations, where a standard critic is prone to overestimating values. 
To obtain more reliable value estimates, we adopt the value-learning objective of Implicit Q-Learning (IQL)~\citep{kostrikov2022IQL} and train a Q-function $Q_\phi$ together with a value function $V_\psi$:
\begin{equation}\label{eq:critic_V}
\mathcal{L}_{V}
=
\mathbb{E}_{(s,\mathbf{a})\sim \mathcal{D}}
\left[
L_2^\tau\!\left(Q_\phi(s,\mathbf{a})-V_\psi(s)\right)
\right],
\end{equation}
\begin{equation}\label{eq:critic_Q}
\mathcal{L}_{Q}
=
\mathbb{E}_{(s,\mathbf{a},s')\sim \mathcal{D}}
\left[
\left(
r(s,\tilde a)+\gamma V_\psi(s')-Q_\phi(s,\mathbf{a})
\right)^2
\right],
\end{equation}
where $L_2^\tau(u)=|\tau-\mathbf{1}(u<0)|u^2$ denotes the expectile regression loss, and $\mathbf{a}=a_{1:H}$ is an action chunk of horizon $H$. 
This design inherits the conservatism of IQL by learning values primarily from actions supported by the offline data, leading to more reliable estimates in failure-prone states.
In our implementation, the environment executes only a single control action obtained by temporally aggregating the chunk, denoted by $\tilde a=g(\mathbf{a})$. The immediate reward and transition are determined by $\tilde a$. 
Accordingly, we treat $Q_\phi(s,\mathbf{a})$ as a chunk-conditioned critic that scores candidate action chunks, while its TD target is defined using the executed action $\tilde a$.

\textbf{Failure Attribution.}
Given a failed trajectory $\xi=\{(s_t,a_t)\}_{t=0}^{T}$, we retrospectively extract action chunks likely responsible for the failure as negative examples.
For each chunk $\mathbf a_t=a_{t:t+H-1}$, we measure its effect by the value change across the chunk:
$
\Delta V_t = V_{\psi}(s_{t+H}) - V_{\psi}(s_t).
$
With negative rewards before task completion, values typically increase as the agent approaches success. A pronounced decrease in value indicates unstable or failure-inducing behavior \citep{sailor, asking_for_help_icra2023}.
Rather than using a fixed threshold, we rank all chunks in the failed trajectory by $\Delta V_t$ and select those at or below the \(\rho\)-th percentile as negative samples:
\begin{equation}\label{eq:delta_V}
\xi^{-} = \left\{(s_t, \mathbf a_t^{-}) \mid \Delta V_t \leq \mathrm{Percentile}_{\rho}(\{\Delta V_j\}) \right\}.
\end{equation}
\begin{wrapfigure}{r}{0.52\textwidth}
\begin{minipage}{0.52\textwidth}
\vspace{-2.5em}
\begin{algorithm}[H]
\caption{{\METHOD} (\method)}
\label{alg:FAR}
\begin{algorithmic}[1]
    \STATE \textbf{Initialize:} pretrained policy $\pi_\theta$, critic $(V_\psi,Q_\phi)$, maximum number of attempts $N$
    \FOR{$i=1$ to $N$}
        \STATE Execute the policy and collect a trajectory $\xi_i$
        \IF{$\textsc{Success}(\xi_i)$}
            \STATE \textbf{break}
        \ENDIF
        \STATE Identify candidate failure-inducing action chunks using Eq.~(\ref{eq:delta_V})
        \STATE Construct contrastive pairs using Eq.~(\ref{eq:filter}) and Eq.~(\ref{eq:action_Q})
        \STATE Update the policy for $n_{\text{tta}}$ steps using Eq.~(\ref{eq:preference})
        \STATE Return the robot to its initial configuration
        \STATE Prepare for the next attempt with the updated policy and action perturbation in Eq.~(\ref{eq:perturbation})
    \ENDFOR
\end{algorithmic}
\end{algorithm}
\vspace{-2em}
\end{minipage}
\end{wrapfigure}
\textbf{Contrastive Sample Construction.}
Learning directly from only negative samples can be unstable and may distort the policy's learned representation. Inspired by contrastive and preference-based learning \citep{dpo, wallace2024diffusion},
we instead construct pairwise supervision using both negative and positive samples, which provides a more informative update direction.
For a state $s_t$ from a failed trajectory and an identified negative chunk $\mathbf a_t^{-}$, we sample a set of candidates from the current policy:
$\mathcal{C}(s_t) = \left\{ \mathbf a_t^{(1)}, \mathbf a_t^{(2)}, \dots, \mathbf a_t^{(M)} \right\}, 
\quad \mathbf a_t^{(i)} \sim \pi_\theta(\cdot \mid s_t).$
To avoid selecting actions that are either too similar to $\mathbf a_t^{-}$ or too far away from the policy distribution, we first filter candidates according to their distance to $\mathbf a_t^{-}$:
\begin{equation}\label{eq:filter}
\mathcal{C}_{\mathrm{safe}}(s_t) =
\left\{
\mathbf a_t^{(i)} \in \mathcal{C}(s_t)
\;\middle|\;
d_{\min} \leq d\!\left(\mathbf a_t^{(i)}, \mathbf a_t^{-}\right) \leq d_{\max}
\right\},
\end{equation}
where $d(\cdot,\cdot)$ denotes the $L^2$ distance between the action chunks, and $d_{\min}, d_{\max}$ define a moderate deviation range. 
We then rank the remaining candidates using the critic $Q_{\phi}(s_t,\mathbf a_t^{(i)})$ and select the top-$K$ highest-valued ones as positive samples:
\begin{equation}\label{eq:action_Q}
\mathcal{A}_t^{+}
=
\operatorname{TopK}_{\mathbf a \in \mathcal{C}_{\mathrm{safe}}(s_t)}
Q_{\phi}(s_t,\mathbf a).
\end{equation}
Together with $\mathbf a_t^{-}$, this forms a set of contrastive preference pairs
$
\{(\mathbf a_t^{+}, \mathbf a_t^{-}) \mid \mathbf a_t^{+}\in \mathcal{A}_t^{+}\},
$
indicating that each positive chunk should be preferred over the negative one.

\textbf{Failure-Contrastive Preference Adaptation.}
Using the constructed contrastive pairs, we adapt the policy with a preference optimization objective \citep{dpo}. For diffusion policies, the action likelihood is not explicitly tractable. Following prior work on preference optimization for diffusion models \citep{wallace2024diffusion}, we use the negative denoising loss as a surrogate preference score. Specifically, for an action chunk $\mathbf a$ conditioned on state $s$, we define
$\ell_{\mathrm{diff}}(\mathbf a,s;\theta)
=
\mathbb{E}_{k,\boldsymbol{\epsilon}}
\left[
\left\|
\boldsymbol{\epsilon}
-
\boldsymbol{\epsilon}_\theta(\mathbf a_k,s,k)
\right\|^2
\right],$
where $\mathbf a_k$ is obtained by adding Gaussian noise to $\mathbf a$ at diffusion step $k$.
We then optimize the pairwise preference objective
\begin{equation}\label{eq:preference}
\mathcal{L}_{\mathrm{pref}}
=
-\log \sigma\left(
\beta
\left[
\ell_{\mathrm{diff}}(\mathbf a_t^{-},s_t;\theta)
-
\ell_{\mathrm{diff}}(\mathbf a_t^{+},s_t;\theta)
\right]
\right),
\end{equation}
where \(\sigma\) is the sigmoid function and \(\beta > 0\) controls the strength of the preference signal. This objective encourages the model to assign lower denoising error to the preferred chunk than to the failed chunk. In this way, safer and higher-value recovery actions become more likely under the adapted policy, while previously failed behaviors are suppressed.
In practice, test-time adaptation only requires 10 gradient steps and can be completed within seconds.

\textbf{Perturbation for Retry Exploration.}
Although FCPA enables learning from previous failures, it remains limited by the support of the offline policy. To improve exploration at out-of-distribution states, we inject lightweight perturbations into the executed action during retry. In simulation, simple Gaussian perturbations are often sufficient. 
For real-world deployment, we further smooth the perturbation over time to reduce high-frequency jitter.
Specifically, with probability $\epsilon_{\text{explore}}$, we sample a target perturbation $\boldsymbol{\delta}^{\star} \sim \mathcal{N}(\mathbf 0,\Sigma)$ and keep it fixed for the next few steps; otherwise, we set $\boldsymbol{\delta}^{\star}=\mathbf 0$. The applied perturbation is updated by exponential smoothing:
$
\boldsymbol{\delta}_t
=
\alpha \boldsymbol{\delta}_{t-1}
+
(1-\alpha)\boldsymbol{\delta}^{\star},
$
where $\alpha\in[0,1)$ is a smoothing coefficient. Let $\mathbf a_t \sim \pi_\theta(\cdot\mid s_t)$ denote the predicted action chunk and $\hat a_t=g(\mathbf a_t)$ the temporally aggregated action. The executed action is
\begin{equation}\label{eq:perturbation}
\tilde a_t=\hat a_t+\boldsymbol{\delta}_t.
\end{equation}
This perturbation provides simple local exploration during retry, while the temporal smoothing helps maintain stable execution on real robots. The overall \method{} is summarized in Algorithm~\ref{alg:FAR}.

\subsection{Continual Policy Improvement}
\label{sec:online}
Failures reveal the boundary of the current policy. When a retry succeeds, the resulting recovery trajectory provides useful supervision for improving robustness. We therefore integrate \method{} into a policy improvement loop, where the robot continually interacts with the environment, performs failure-aware retry, and uses both direct-success and successful-recovery trajectories for policy improvement.
We maintain three replay buffers:
$\mathcal{D}_{\mathrm{exp}}$, $\mathcal{D}_{\mathrm{succ}}$, and $\mathcal{D}_{\mathrm{fail}}$,
which store offline expert demonstrations, successful online trajectories (including recovered ones), and failure trajectories, respectively. 

\textbf{Critic Learning.}
Since online rollouts can be suboptimal and may contain stagnation or partial failures, naively cloning all collected data can degrade the policy. We therefore leverage the critic to estimate the quality of state-action pairs and reweight the policy update accordingly. Specifically, we adopt an IQL-style actor-critic framework. Given the aggregated replay buffer
$\mathcal{D} = \mathcal{D}_{\mathrm{exp}} \cup \mathcal{D}_{\mathrm{succ}} \cup \mathcal{D}_{\mathrm{fail}},$
we first optimize $V_\psi$ and $Q_\phi$ with Eq. (\ref{eq:critic_V}) and Eq. (\ref{eq:critic_Q}).

\textbf{Advantage-Weighted Policy Update.}
After critic training, we compute the advantage of each sample as
$A(s,\mathbf{a})=Q_\phi(s,\mathbf{a})-V_\psi(s),$ 
and assign it the weight
$w(s,\mathbf{a})=\exp\!\left(\frac{A(s,\mathbf{a})}{\eta}\right),$
where $\eta$ is a temperature parameter. To avoid learning poor behaviors, the actor is updated only on
$\mathcal{D}_{\mathrm{exp}}\cup\mathcal{D}_{\mathrm{succ}}$.
As the policy is parameterized as a diffusion policy, we use an advantage-weighted denoising objective. Let $\mathbf{a}_0$ denote the ground-truth action chunk, $\mathbf{a}_k$ the noisy sample at diffusion step $k$, and $\epsilon_\theta(\mathbf{a}_k,s,k)$ the predicted noise. The actor objective is
\begin{equation}\label{eq:awr}
\mathcal{L}_{\pi} =
\mathbb{E}_{(s,\mathbf{a}_0)\sim \mathcal{D}_{\mathrm{exp}}\cup\mathcal{D}_{\mathrm{succ}},\, k,\, \epsilon}
\left[
w(s,\mathbf{a}_0)
\left\|
\epsilon-\epsilon_\theta(\mathbf{a}_k,s,k)
\right\|^2
\right].
\end{equation}
This training loop gradually expands the policy support while prioritizing high-value behaviors. In particular, successful recovery trajectories provide supervision on hard states where the initial policy fails, improving both policy robustness and value estimation over time.

\section{Experiments}

\subsection{Experimental Setup}
\textbf{Settings.}
For each task, we assume access to an offline demonstration dataset.
Using this dataset, we train a diffusion policy with behavior cloning and the corresponding critic models. The policy takes visual observations and robot proprioception as input. Through experiments, we aim to answer the following research questions: (1) Can {\method} improve evaluation performance over the pretrained policy? (2) Can the policy learn from collected recovery experience and continually improve over time? (3) Can {\method} reduce environment reset costs and improve the data efficiency of online learning? Our experiments are conducted on a single NVIDIA A5000 GPU. Additional implementation details are provided in the Appendix (Section~\ref{appendix:implementation}). 

\textbf{Tasks.}
We evaluate the proposed method in the ManiSkill~\citep{taomaniskill3}, RoboSuite~\citep{robosuite2020}, and RoboMimic~\citep{robomimic2021} environments. Our simulated evaluation covers nine vision-based manipulation tasks with different levels of difficulty and task horizons. For each task, we run 50 evaluation episodes. Unless otherwise specified, the maximum number of attempts per episode is set to 5. We report the average success rate and standard deviation over three random seeds. 
We also evaluate our method on three real-world manipulation tasks using a 7-DoF xArm. For each task, we run 20 evaluation episodes, with the maximum number of attempts per episode set to 3. The task setup is shown in Figure~\ref{fig:setup}.

\textbf{Baselines.}
(1) \textbf{DP (Base policy)}: Diffusion Policy \citep{chi2023diffusionpolicy} with standard policy rollout during evaluation. 
(2) \textbf{DP-NR (Naive Retry)}: The same diffusion policy is given multiple attempts to complete the task, without model updates or environment resets. 
(3) \textbf{DP-BGR (Bellman-Guided Retrials)}: \citet{du2024err} proposed a non-parametric method based on rejection sampling to avoid repeated failures during retries. We implement a compatible version in our setting for comparison. 
(4) \textbf{DP-FAR (Ours)}: Diffusion Policy with the proposed {\METHOD}. 
In the policy improvement experiments, we add simple Gaussian action noise to all baselines during data collection to encourage exploration.

\subsection{Test-Time Failure Recovery}
\input{table/main}

\textbf{Simulated Evaluation.}
As shown in Table~\ref{tab:tta_maniskill} and Table~\ref{tab:tta_robomimic}, allowing retries improves evaluation success rates, which highlights the importance of non-reset retries for improving policy robustness. 
However, DP-NR often repeats similar behaviors across attempts, which limits its effectiveness.
For the non-parametric method DP-BGR, a key challenge is that previous failed attempts may alter the environment, leading to distribution shifts in the observed state and reducing the usefulness of failure feedback from earlier attempts.
In contrast, the proposed FAR performs a local test-time update based on previous failures. Requiring only a few seconds for 10 optimization steps, {\method} improves performance across most tasks and with different numbers of training demonstrations.

\textbf{Real-World Evaluation.}
Similar trends are observed in real-world tasks. 
As shown in Figure~\ref{fig:real}, on shorter-horizon tasks such as Drawer, DP performs well, and all methods achieve high success rates. On more challenging tasks such as Pot and Tea, however, recovery becomes substantially harder, as the policy must identify which actions led to failure and then explore alternative behaviors. By combining test-time adaptation from previous failures with perturbation-based exploration beyond the training distribution, {\method} achieves higher recovery success rates in these challenging scenarios.

\begin{table*}[t]
\begin{minipage}[b]{.49\textwidth}
\centering
\includegraphics[width=\textwidth]{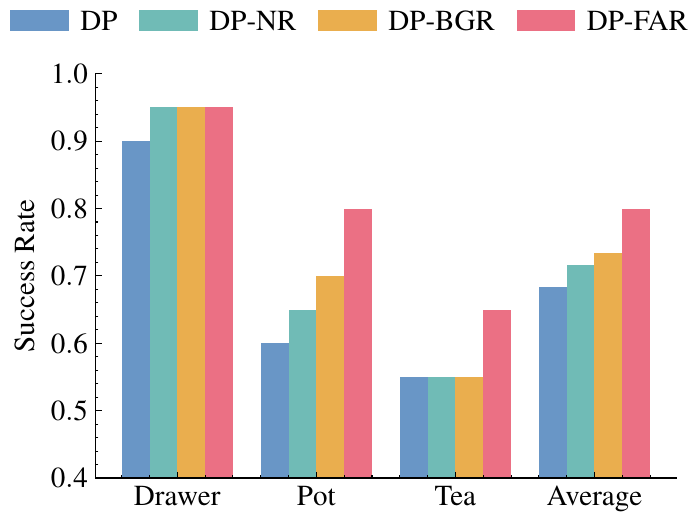}
\captionof{figure}{\textbf{Comparison Across Real-World Tasks.} We conduct experiments on three real-world manipulation tasks that evaluate pushing, pick-and-place, and pouring skills.}
\label{fig:real}
\end{minipage}
\hfill
\begin{minipage}[b]{.49\textwidth}
\centering
\includegraphics[width=\textwidth]{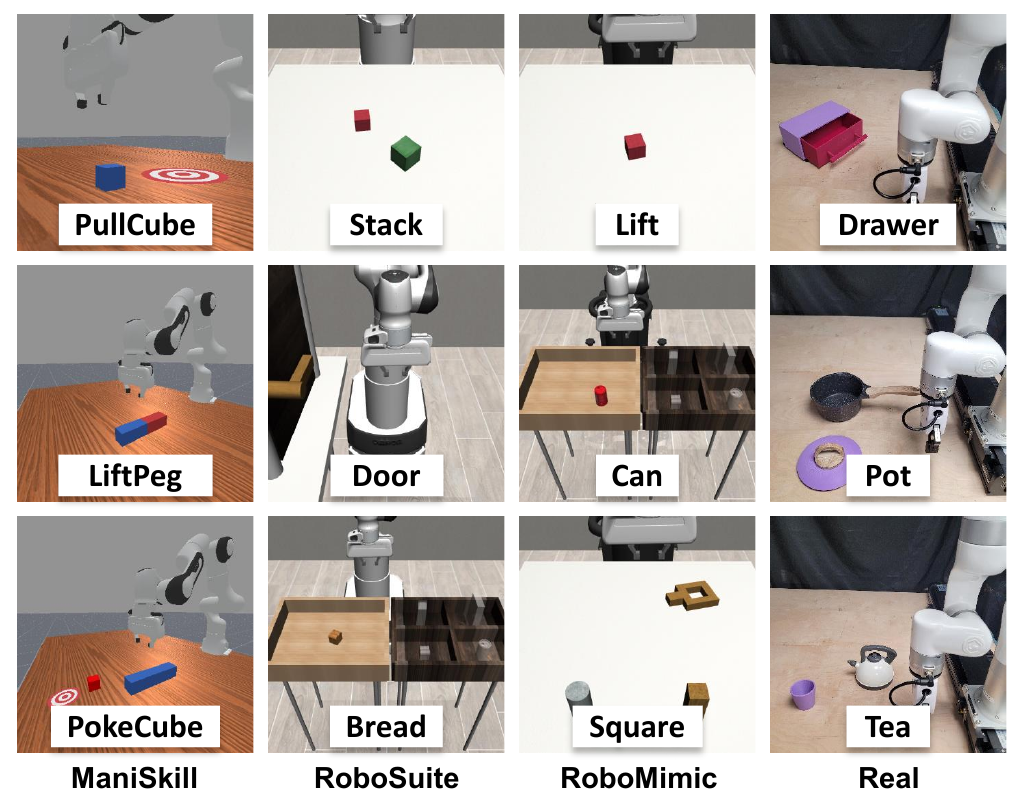}
\captionof{figure}{\textbf{Task Setup.} We evaluate {\method} in three simulated environments covering nine manipulation tasks, as well as three real-world tasks on a 7-DoF xArm platform.}
\label{fig:setup}
\end{minipage}
\end{table*}

\subsection{Continual Policy Improvement}

\begin{figure*}[t]
  \vspace{-0.15in}
  \centering
    {\label{IQL:legend}{\includegraphics[width=0.7\linewidth]{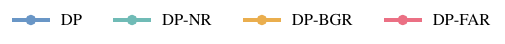}}
    }

  {
  \label{IQL:subfig1}
  \includegraphics[height=3.2cm]{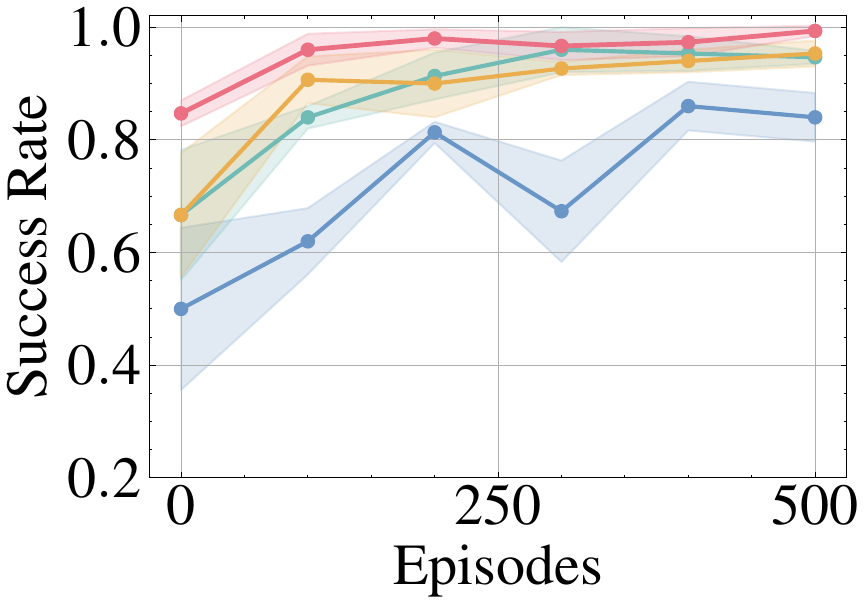}
  }\hfill
  {
    \label{IQL:subfig2}\includegraphics[height=3.2cm]{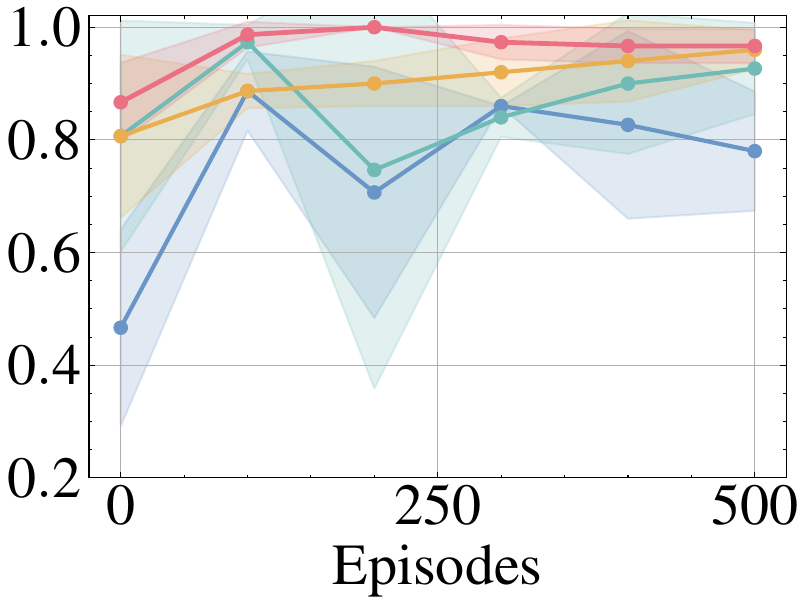}
  }\hfill
  {
\label{IQL:subfig3}\includegraphics[height=3.2cm]{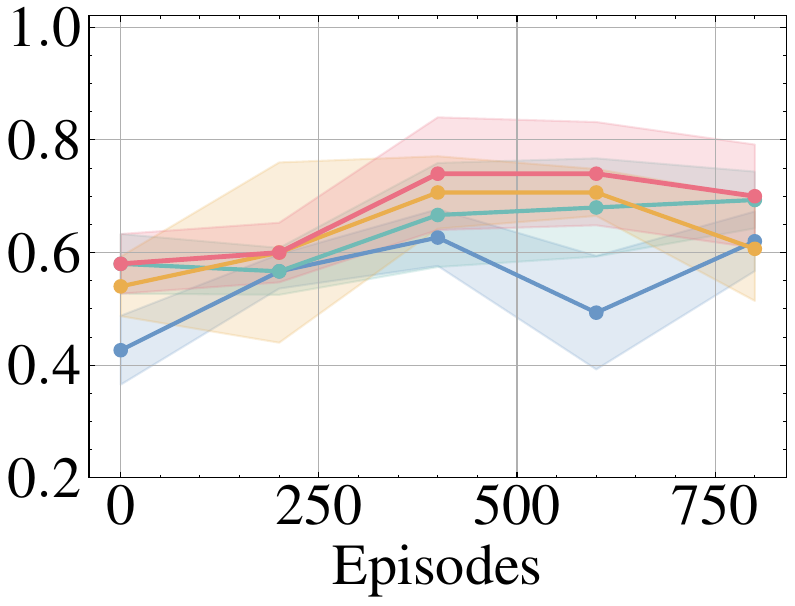}
  }

    \subfloat[Lift $\left | \mathcal{D}_{\mathrm{exp}} \right |=5 $]{\label{IQL-step:subfig1}\includegraphics[height=3.2cm]{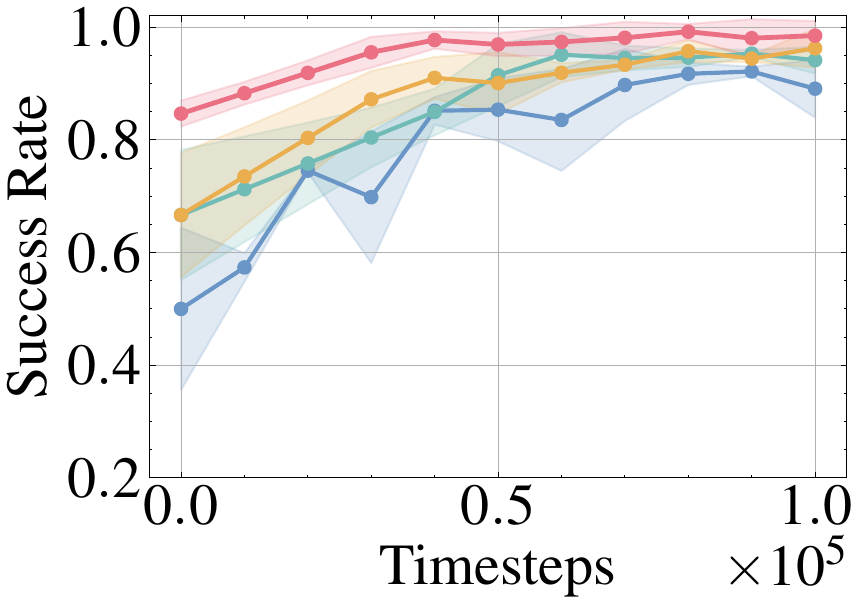}
  }\hfill
   \subfloat[Door $\left | \mathcal{D}_{\mathrm{exp}} \right |=3 $]
  {
    \label{IQL-step:subfig2}\includegraphics[height=3.2cm]{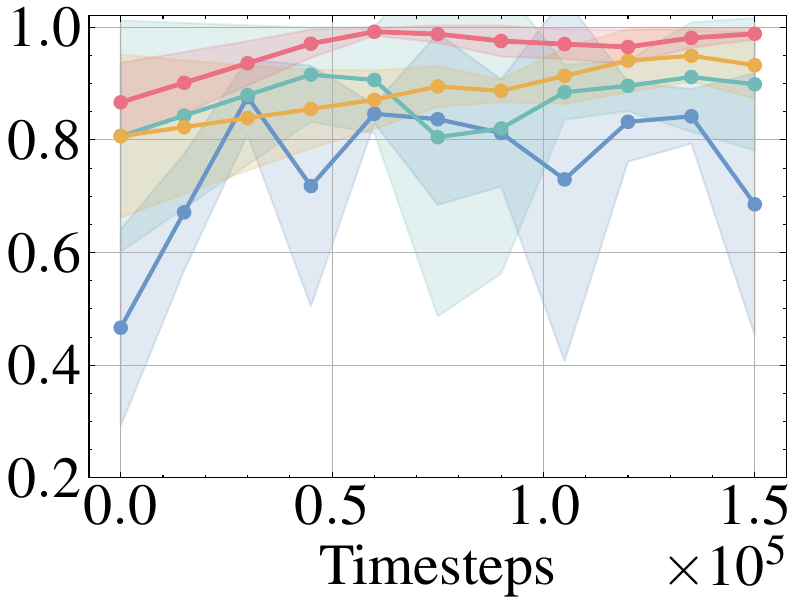}
  }\hfill
   \subfloat[Can $\left | \mathcal{D}_{\mathrm{exp}} \right |=10 $]
  {
\label{IQL-step:subfig3}\includegraphics[height=3.2cm]{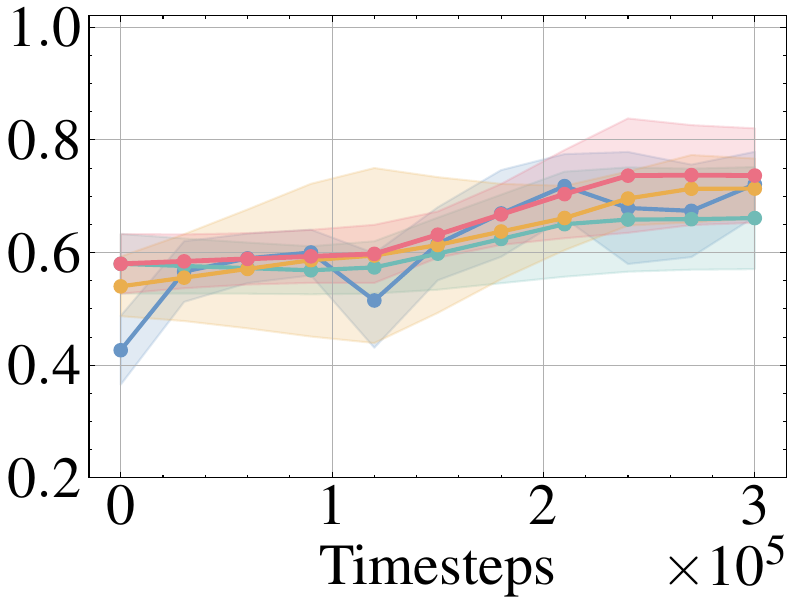}
  }
  \vspace{-0.0in}
  \caption{\textbf{Results on Continual Policy Improvement.}
  {\method} improves performance through online interactions, while increasing data efficiency and reducing the number of costly environment resets.
  }
  \label{fig:online-IQL}
\vspace{-4mm}
\end{figure*}

As shown in Figure~\ref{fig:online-IQL}, the policy steadily improves with additional environment interactions and newly collected experience. Data efficiency remains a key challenge in online learning. From the perspective of environment resets, allowing more attempts within each episode generally yields more training data, resulting in more stable learning signals and more robust policy improvement. 
Among all methods, {\method} achieves the best performance and data efficiency by enabling more informative retries and exploration after failure. This helps the robot learn from hard negative examples and expand its capability boundary, providing more valuable data for online learning. \method{} also achieves greater improvement per environment timestep by focusing retry interactions on challenging failure cases near the current policy's capability boundary. More results are shown in the Appendix (Section~\ref{appendix:result}). 

\subsection{Ablation Studies}
\input{table/ablation}

\begin{wrapfigure}{r}{0.55\textwidth}
\begin{minipage}{0.55\textwidth}
  \vspace{-0.25in}
  \centering
  \subfloat[Door $\left | \mathcal{D}_{\mathrm{exp}} \right |=3 $]
  {
\label{ablation:subfig1}\includegraphics[
height=2.7cm
]{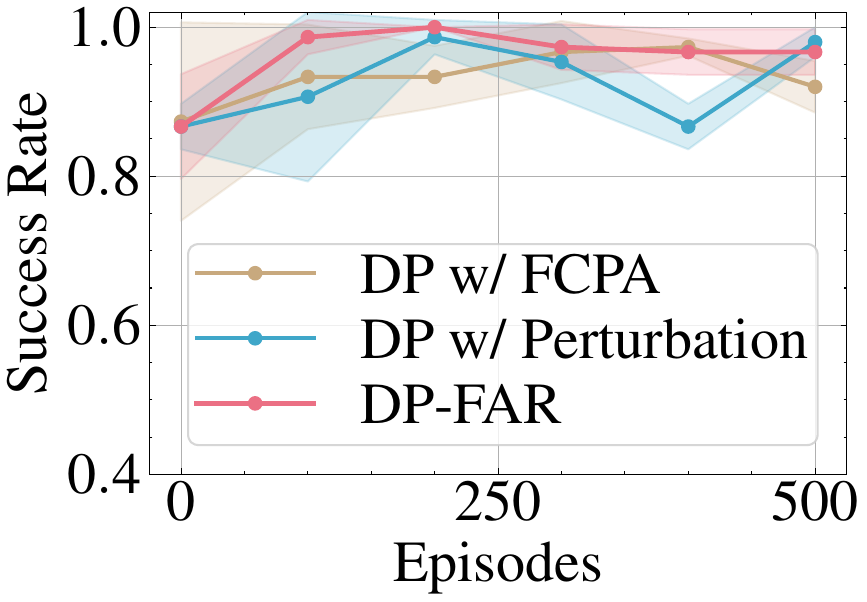}
  }
   \subfloat[Square $\left | \mathcal{D}_{\mathrm{exp}} \right |=100 $]
  {
\label{ablation:subfig2}\includegraphics[
height=2.7cm
]{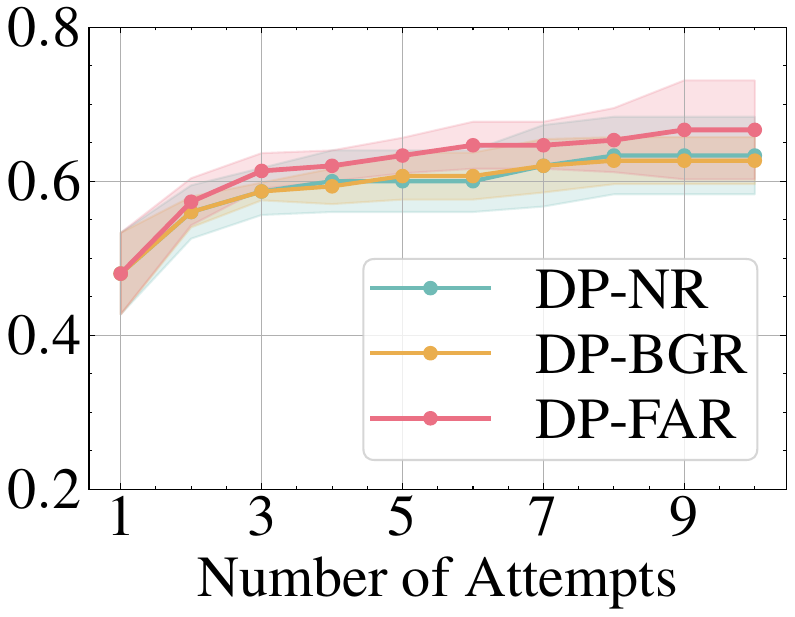}
  }
  \vspace{-1mm}
  \caption{
  (a) {\method} benefits from both failure adaptation and local exploration. (b) {\method} improves the success rate across different numbers of attempts.}
  \label{fig:ablation}
  \vspace{-12mm}
\end{minipage}
\end{wrapfigure}

\textbf{Design of the Retry Algorithm.} We study the contribution of each component in {\method}, and present results in Table~\ref{tab:tta_ablation}. Compared with naive retry, both FCPA and action perturbation improve performance on average. FCPA helps the policy avoid repeating previous failures and make more effective retries, which is especially beneficial on longer-horizon tasks such as Door. Perturbation encourages the policy to try novel actions, allowing it to reach new states and escape local optima. By combining both components, FAR achieves the best overall performance. Similar trends are observed during online learning, as shown in Figure~\ref{ablation:subfig1}.

\textbf{Failure Attribution.} Accurate credit assignment is crucial for long-horizon decision-making tasks. Without the proposed value-based negative transition detection, adapting on the entire trajectory degrades performance, since not all transitions contribute to failure. 
Compared with a fixed threshold, we empirically find that using a percentile of value differences yields better overall performance by providing more stable training samples and gradients for test-time adaptation.

\textbf{Adaptation Objective.} We evaluate the roles of positive and negative samples in FCPA. Without positive samples, the policy only learns to avoid previous actions, which can quickly distort the original score structure of the policy. Without negative samples, adaptation degenerates into self-imitation and cannot effectively exploit failure feedback, resulting in weaker retry performance. We also compare FCPA with a standard RL objective, advantage-weighted regression (AWR)~\cite{peng2019advantage}. AWR performs worse in our setting because it mainly exploits positive updates from failure trajectories, which does not provide sufficient signal to explicitly avoid failed behavior. 

\textbf{Number of Attempts.}
We also report success rates for different maximum numbers of attempts per evaluation episode. 
As shown in Figure~\ref{ablation:subfig2}, the success rate increases rapidly with more attempts at first and then gradually saturates, ultimately limited by the model's initial capability.

\textbf{Additional Ablations.} We present additional results in the Appendix (Section~\ref{appendix:result}).

\newpage
\subsection{Qualitative Results}
\begin{figure*}[t]
  \begin{center}
    \centerline{\includegraphics[width=0.87\linewidth]{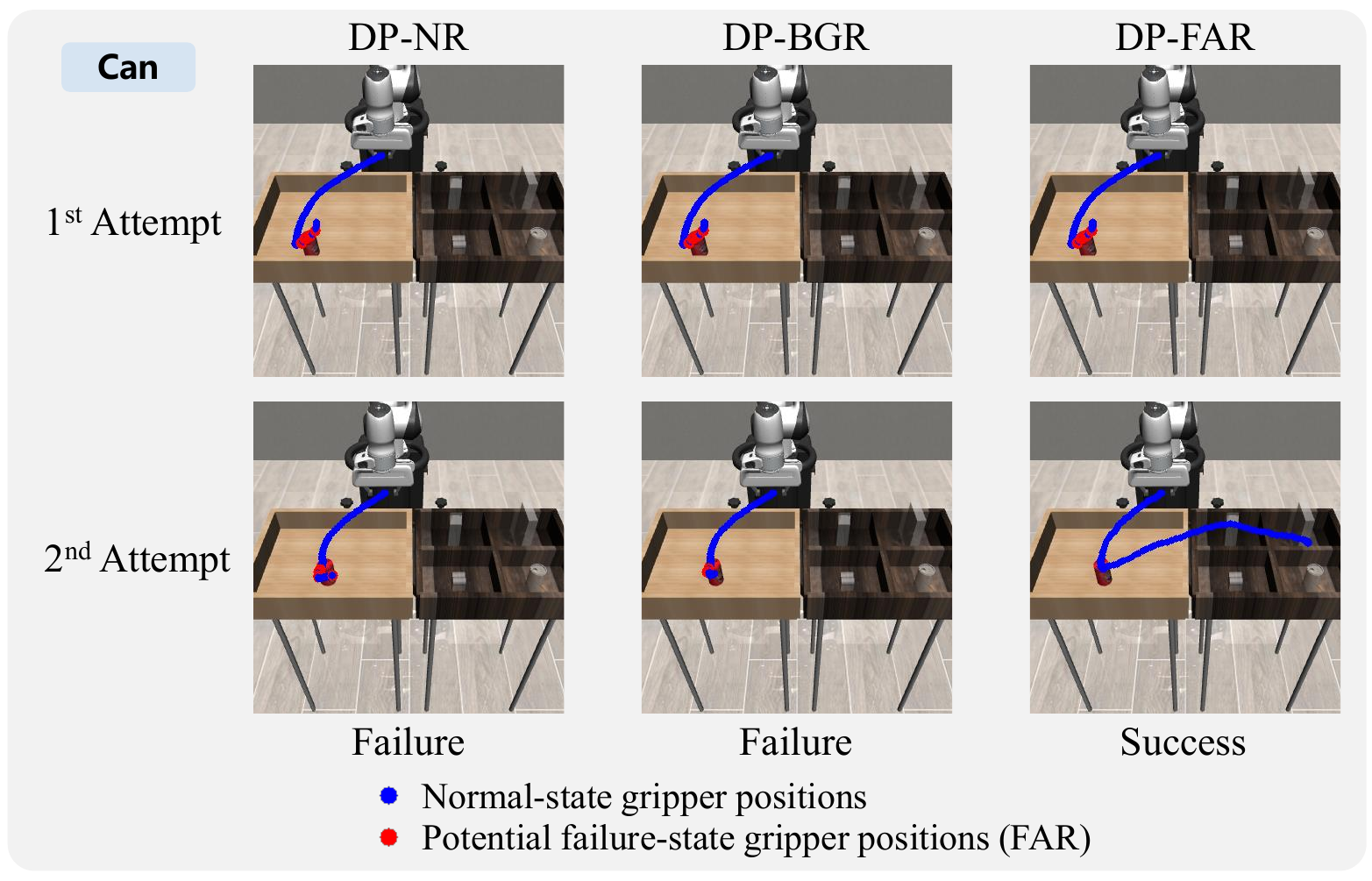}}
    \caption{
    \textbf{Qualitative Results on Can.} We visualize the gripper-center trajectory, where red and blue points indicate potentially failure-inducing and normal positions identified by FAR.
    }
    \vspace{-8mm}
    \label{fig:qualitative}
  \end{center}
\end{figure*}

In Figure~\ref{fig:qualitative}, we visualize the gripper-center trajectory, where red and blue points indicate gripper positions at potentially failure-prone and normal states identified by FAR, respectively. 
FAR localizes the failure to the contact phase, where the gripper pushes the can away, and updates the policy to avoid the same mistake, successfully grasping the can on the second attempt. In contrast, BGR's non-parametric correction is sensitive to changes in the environment state, as repulsion from the previous failed action may no longer provide effective guidance once the failed interaction displaces the can.

\section{Conclusion}
\label{sec:conclusion}
We propose {\method}, a method that enables a policy to learn from failures and improve at test time. We further show that the collected recovery experience can be incorporated into continual policy improvement, helping the policy address hard failure cases and improve over time. We evaluate {\method} in three simulation benchmarks across nine manipulation tasks, as well as on three real-world tasks. The results show that {\method} improves the performance of pretrained policies and enables recovery without environment resets. Our results on continual policy improvement further show that {\method} increases learning efficiency and reduces the need for human intervention, demonstrating its potential for robust and data-efficient self-improvement during deployment.

\textbf{Limitations and Future Work.} First, our experiments focus on diffusion policies trained from scratch in single-task settings. An important future direction is to evaluate the proposed method on large-scale pretrained policies, such as VLA models, which may offer richer prior knowledge and broader generalization. Second, our current framework does not include a principled failure detection mechanism, and instead relies on environment feedback and time limits. In future work, we plan to incorporate pretrained models, such as VLMs, for task tracking and failure detection. 
In addition, further efforts are needed to improve the safety of exploration during recovery.



	

\acknowledgments{
This work was funded in part by gifts from Stack AV and Skylark Labs. We would like to thank Yiqi Wang for helpful discussions on the experimental setup.
}


\bibliography{main}  

\clearpage
\appendix
\input{appendix}

\end{document}

%% file: table/main.tex
\begin{table*}[t]
  \centering
  \caption{\textbf{Results on ManiSkill and RoboSuite Benchmarks.} Each task is evaluated over 50 trials, and we report the mean success rate and standard deviation across 3 random seeds. {\method} consistently outperforms all baselines, achieving an average improvement of 16.4\% over the base policy.
  }
  \label{tab:tta_maniskill}
  \vspace{-4mm}
  \begin{center}
  \setlength{\tabcolsep}{0.75mm}
  \begin{tabular}{l *{7}{S[table-format=2.1(2.1)]}}
    \toprule
    Benchmark & \multicolumn{3}{c}{ManiSkill} & \multicolumn{3}{c}{RoboSuite} \\
    \cmidrule(lr){2-4}\cmidrule(lr){5-7}
    Task & {LiftPeg} & {PokeCube} & {PullCube} & {Door} & {Bread} & {Stack} & {Average} \\
    \cmidrule(lr){2-2}\cmidrule(lr){3-3}\cmidrule(lr){4-4}\cmidrule(lr){5-5}\cmidrule(lr){6-6}\cmidrule(lr){7-7}\cmidrule(lr){8-8}
    
    $|\mathcal{D}_{\mathrm{exp}}|$ & {\phantom{0}20} & {\phantom{0}20} & {\phantom{0}20} & {\phantom{0}3} & {\phantom{0}15} & {\phantom{0}15} & {\text{--}} \\
    \midrule
    DP     & 53.3\pm10.3 & 36.0\pm3.5 & 58.7\pm4.2 & 46.7\pm17.5 & 44.7\pm8.1  & 42.7\pm8.3  & 47.0\pm6.6 \\
    DP-NR  & 57.3\pm8.3  & 37.3\pm4.6 & 64.0\pm5.3 & 80.7\pm20.5 & 54.0\pm11.1 & 61.3\pm16.0 & 59.1\pm8.0 \\
    DP-BGR & 53.3\pm8.1  & 36.0\pm7.2 & 62.7\pm4.2 & 80.7\pm14.5 & 50.7\pm12.2 & 63.3\pm21.0 & 57.8\pm7.7 \\
    \textbf{DP-FAR}  & \bf 58.7\pm6.1 & \bf 39.3\pm8.1 & \bf 70.7\pm5.0  & \bf 86.7\pm7.0  & \bf 56.7\pm1.2  & \bf 68.7\pm13.3  & \bf 63.4\pm4.5 \\
    \bottomrule
  \end{tabular}
  \end{center}
  \vspace{-1mm}
  \caption{\textbf{Results on the RoboMimic Benchmark.}
    Across different numbers of training demonstrations, {\method} improves the success rate in most settings.
  }
  \label{tab:tta_robomimic}
  \vspace{-4mm}

  \begin{center}
  \setlength{\tabcolsep}{0.95mm}
  \begin{tabular}{l *{7}{S[table-format=2.1(2.1)]}}
    \toprule
    Task   & \multicolumn{2}{c}{Lift}& \multicolumn{2}{c}{Can} & \multicolumn{2}{c}{Square}  & {Average} \\
    \cmidrule(lr){2-3}\cmidrule(lr){4-5}\cmidrule(lr){6-7}
        
    $|\mathcal{D}_{\mathrm{exp}}|$ & {\phantom{0}5} & {\phantom{0}10} & {\phantom{0}10} & {\phantom{0}20} & {\phantom{0}50} & {\phantom{0}100} & {\text{--}} \\
    \midrule
    DP     & 50.0\pm14.4 & 70.7\pm2.3  & 42.7\pm6.1  & 82.7\pm3.1  & 19.3\pm4.2  & 48.0\pm5.3  & 52.2\pm4.8 \\
    DP-NR  & 66.7\pm11.5  & 87.3\pm10.3  & 54.7\pm7.0  & 86.0\pm2.0  & 34.7\pm9.9  & 60.0\pm4.0  & 64.9\pm1.0 \\
    DP-BGR  & 66.7\pm11.0  & 84.7\pm7.6  & 54.0\pm5.3  & \bf 90.7\pm1.2  & 34.0\pm8.7  & 60.7\pm3.1  & 65.1\pm1.3 \\
    \textbf{DP-FAR}  & \bf 84.7\pm2.3  & \bf 92.0\pm2.0  & \bf 58.0\pm5.3  & 89.3\pm6.1  & \bf 38.0\pm11.1  & \bf 63.3\pm2.3  & \bf 70.9\pm3.0 \\
    \bottomrule
  \end{tabular}
  \end{center}
  \vskip -0.1in
\end{table*}

%% file: table/ablation.tex
\begin{table*}[t]
\centering
  \caption{\textbf{Results of Ablation Studies.} We conduct ablation studies to examine the design choices of {\method}. Specifically, we compare different components of the retry algorithm, failure attribution methods, and adaptation objectives. The \colorbox{lightgray}{gray rows} indicate the default settings in each ablation. 
  } 
  \vspace{-4mm}
  \label{tab:tta_ablation}
  \begin{center}
  \small
      \setlength{\tabcolsep}{0.8mm}
        \begin{tabular}{l *{7}{S[table-format=2.1(2.1)]}}
          \toprule
          Task  & {PullCube} & {LiftPeg} & {Bread} & {Door}  & \multicolumn{2}{c}{Lift} &  {Average} \\
          \cmidrule(lr){2-2}\cmidrule(lr){3-3}\cmidrule(lr){4-4}\cmidrule(lr){5-5}\cmidrule(lr){6-7}
    $|\mathcal{D}_{\mathrm{exp}}|$ & {\phantom{0}20} & {\phantom{0}20} & {\phantom{0}15} & {\phantom{0}3} & {\phantom{0}5} & {\phantom{0}10} & {\text{--}} \\
          \midrule
          \multicolumn{2}{l}{\textit{Retry Algorithm}} \\          
          DP-NR & 64.0\pm5.3  & 57.3\pm8.3 & 54.0\pm11.1 & 80.7\pm20.5 & 66.7\pm11.5 & 87.3\pm10.3 & 68.3\pm6.2 \\
          DP w/ FCPA & 68.0\pm3.5  & 57.3\pm8.3 & \bf 56.7\pm7.0 & \bf 87.3\pm13.3 & 80.7\pm1.2 & 86.7\pm7.6 & 72.8\pm4.3 \\
          DP w/ Perturbation & 68.7\pm5.0  & 56.7\pm11.0 & 55.3\pm4.2 & 86.7\pm3.1 & 78.0\pm5.3 & 86.7\pm3.1 & 72.0\pm2.3 \\
          \rowcolor{lightgray}DP-FAR & \bf 70.7\pm5.0  & \bf 58.7\pm6.1 & \bf 56.7\pm1.2 & 86.7\pm7.0 & \bf 84.7\pm2.3 & \bf 92.0\pm2.0 & \bf 74.9\pm2.2 \\
          \midrule
          \multicolumn{4}{l}{\textit{Failure Attribution}} \\
          \rowcolor{lightgray}Value-Percentile & \bf 70.7\pm5.0  & \bf 58.7\pm6.1 & 56.7\pm1.2 & \bf 86.7\pm7.0 & \bf 84.7\pm2.3 & 92.0\pm2.0 & \bf 74.9\pm2.2 \\
          Value-Threshold  & 68.0\pm5.3  & 58.0\pm3.5 & \bf 62.7\pm5.0 & 76.7\pm8.3 & 80.0\pm5.3 & \bf 92.7\pm5.8 & 73.0\pm3.6 \\
          Whole Trajectory  & 67.3\pm7.0  & 57.3\pm5.0 & 56.0\pm5.3 & 78.0\pm10.0 & 80.7\pm11.4 & \bf 92.7\pm3.1 & 72.0\pm1.5 \\
          \midrule
          \multicolumn{4}{l}{\textit{Adaptation Objective}} \\       
          \rowcolor{lightgray}Pairwise Preference  & \bf 70.7\pm5.0  & \bf 58.7\pm6.1 & \bf 56.7\pm1.2 & \bf 86.7\pm7.0 & 84.7\pm2.3 & \bf 92.0\pm2.0 & \bf 74.9\pm2.2 \\
          w/o Positive  & 69.3\pm5.0  & 55.3\pm5.0 & 46.0\pm7.2 & 44.0\pm22.0 & 72.7\pm5.0 & 80.7\pm6.1 & 61.3\pm5.5 \\
          w/o Negative  & 68.0\pm6.0  & 58.0\pm9.2 & 53.3\pm4.2 & 81.3\pm10.3 & \bf 86.7\pm5.0 & 90.0\pm0.0 & 72.9\pm1.6 \\
          AWR  & 70.0\pm3.5  & 56.7\pm11.0 & 42.0\pm10.4 & 78.0\pm6.0 & 74.7\pm7.6 & 90.7\pm4.2 & 68.7\pm2.5 \\
          \bottomrule
        \end{tabular}
  \end{center}
  \vskip -0.1in
\end{table*}

%% file: appendix.tex
\section{Appendix Overview}
\begin{itemize}[leftmargin=18 pt, itemsep=3 pt,topsep=1 pt]
\item Section \ref{appendix:result}: Additional experimental results.
\item Section \ref{appendix:implementation}: Implementation details.
\item Section \ref{appendix:expetiment}: Additional experimental details.
\item Section \ref{appendix:related}: Extended related work.
\item Section \ref{appendix:qualitative}: Additional qualitative results.
\end{itemize}


\section{Additional Experimental Results}
\label{appendix:result}

\subsection{Continual Policy Improvement}
\begin{figure*}[th]
  \centering
  {\label{appendix:IQL:legend}{\includegraphics[width=0.7\linewidth]{figure/online/legend.pdf}}
}

    {\label{appendix:IQL:subfig1}\includegraphics[height=3.2cm]{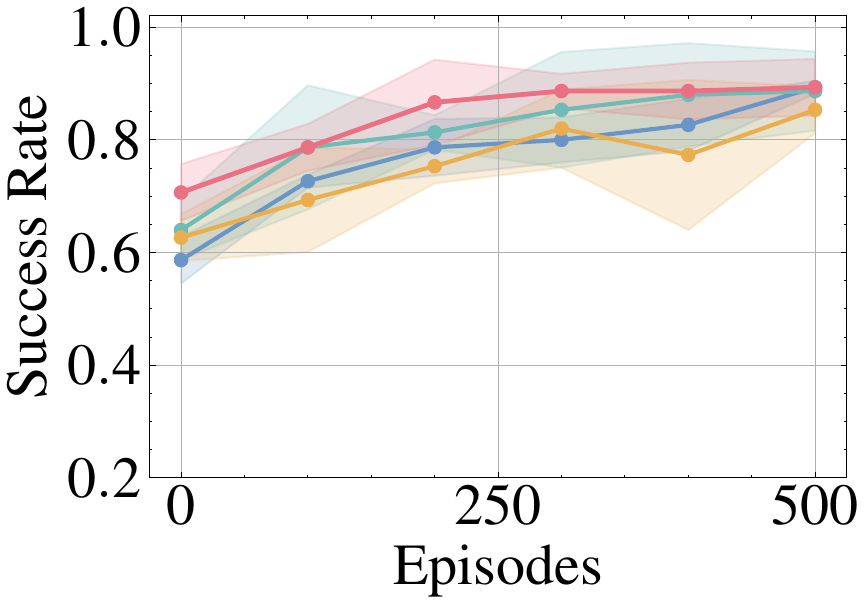}
  }\hfill
    {\label{appendix:IQL:subfig2}\includegraphics[height=3.2cm]{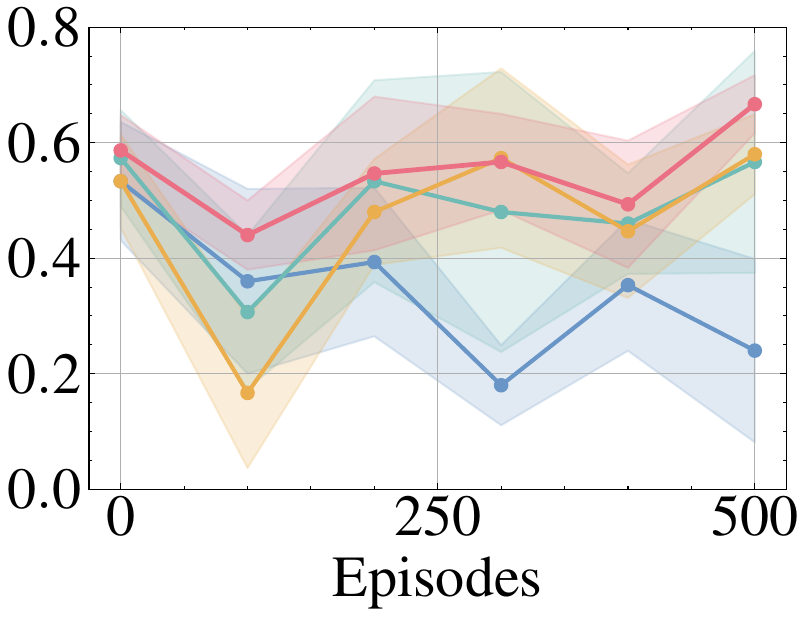}
  }\hfill
   {\label{appendix:IQL:subfig3}\includegraphics[height=3.2cm]{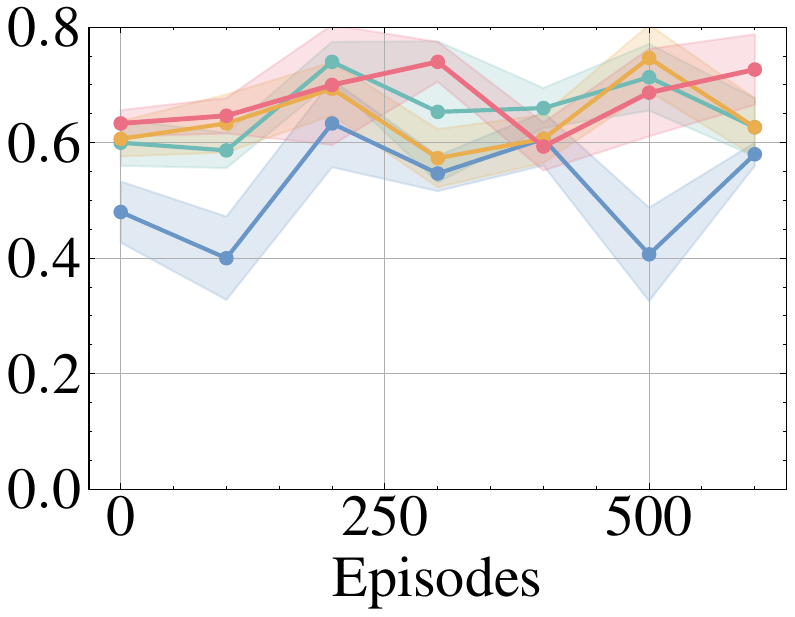}
  }
  

   \subfloat[PullCube $\left | \mathcal{D}_{\mathrm{exp}} \right |=20 $]{\label{appendix:IQL-step:subfig4}\includegraphics[height=3.2cm]{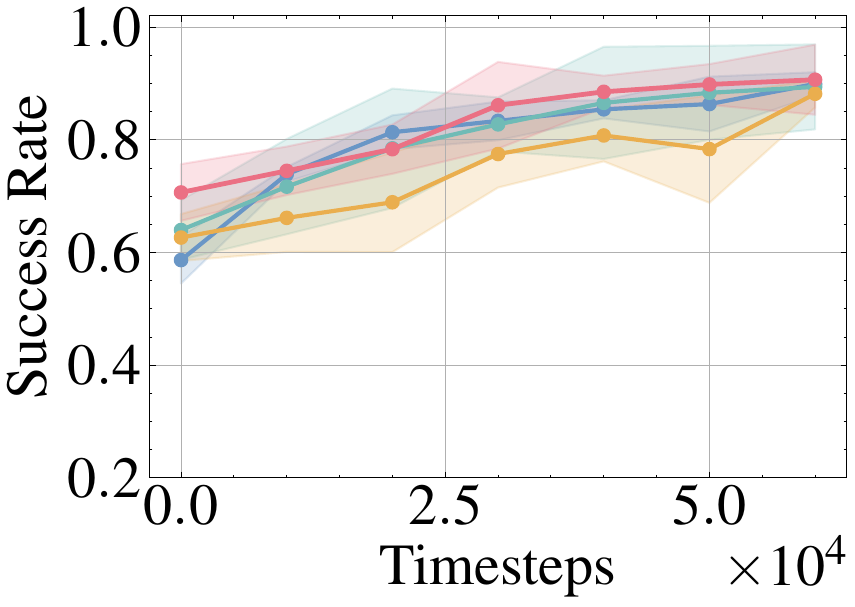}
  }\hfill
   \subfloat[LiftPeg $\left | \mathcal{D}_{\mathrm{exp}} \right |=20 $]{\label{appendix:IQL-step:subfig5}\includegraphics[height=3.2cm]{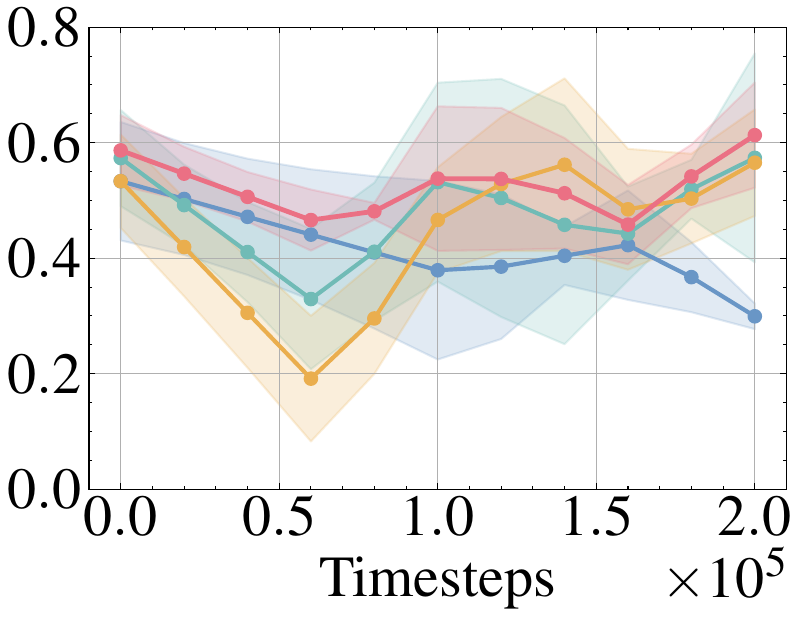}
  }\hfill
   \subfloat[Square $\left | \mathcal{D}_{\mathrm{exp}} \right |=100 $]{\label{appendix:IQL-step:subfig6}\includegraphics[height=3.2cm]{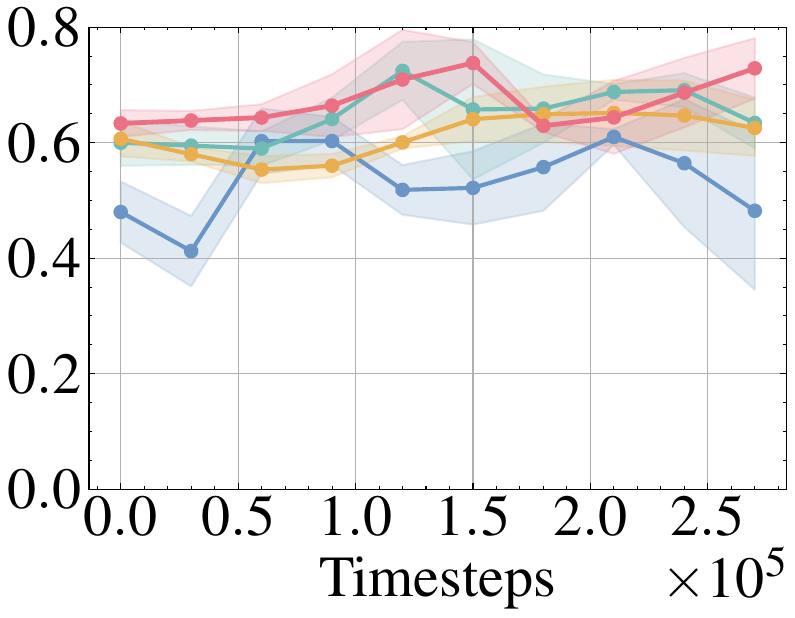}
  }

  \caption{\textbf{Results on Continual Policy Improvement.} The first row reports success rate against the number of online episodes, while the second row reports success rate against environment timesteps.
  {\method} improves performance through online interactions, increasing data efficiency in terms of both resets and timesteps.}
  \label{appendix:fig:online-IQL}
\end{figure*}

In Figure~\ref{appendix:fig:online-IQL}, we present additional results on continual policy improvement. {\method} consistently improves the base policy across different tasks. We also report success rates as a function of environment timesteps. Under the same timestep budget, {\method} achieves higher success rates than the base policy. Although allowing multiple attempts within each episode introduces additional interactions, these interactions are concentrated on challenging failure cases. Such cases are particularly informative, as they often lie near the boundary of the current policy's capabilities, and allocating more timesteps to them leads to greater policy improvement.



\subsection{Additional Comparisons}
\begin{table*}[t]
\centering
\small
  \caption{\textbf{Additional Comparisons.} We evaluate two additional baselines: CCIL~\cite{ke2024ccil} and V-GPS~\cite{nakamoto2024steering}, which represent policy correction and inference-time steering methods, respectively. We also evaluate \method{} with real-time failure detection, denoted as \method{}-Realtime.
}
  \vspace{-4mm}
  \label{appendix:tab:baseline}
  \begin{center}
      \setlength{\tabcolsep}{0.8mm}
        \begin{tabular}{l *{7}{S[table-format=2.1(2.1)]}}
          \toprule
          Task  & {PullCube} & {LiftPeg} & {Bread} & {Door}  & \multicolumn{2}{c}{Lift} &  {Average} \\
          \cmidrule(lr){2-2}\cmidrule(lr){3-3}\cmidrule(lr){4-4}\cmidrule(lr){5-5}\cmidrule(lr){6-7}
    $|\mathcal{D}_{\mathrm{exp}}|$ & {\phantom{0}20} & {\phantom{0}20} & {\phantom{0}15} & {\phantom{0}3} & {\phantom{0}5} & {\phantom{0}10} & {\text{--}} \\
          \midrule
          CCIL \cite{ke2024ccil} & 64.7\pm4.2  & \bf 59.3\pm9.2 & 42.7\pm3.1 & 58.7\pm4.2 & 69.3\pm9.2 & 86.7\pm9.0 & 63.6\pm1.3  \\
          V-GPS \cite{nakamoto2024steering} & 62.0\pm5.3  & 55.3\pm11.0 & 47.3\pm5.8 & \bf 87.3\pm20.2 & 73.3\pm7.0 & 87.3\pm5.8 & 68.8\pm3.5\\
          DP-FAR-Realtime & 62.7\pm2.3  & 46.0\pm5.3 & 40.0\pm3.5 & 42.7\pm4.6 & 61.3\pm3.1 & 84.7\pm8.3 & 56.2\pm0.4 \\
          DP-FAR & \bf 70.7\pm5.0  & 58.7\pm6.1 & \bf 56.7\pm1.2 & 86.7\pm7.0 & \bf 84.7\pm2.3 & \bf 92.0\pm2.0 & \bf 74.9\pm2.2 \\
          \bottomrule
        \end{tabular}
  \end{center}
\end{table*}

We evaluate two additional baselines: CCIL \cite{ke2024ccil}, which uses a learned dynamics model to generate corrective actions for adaptation, and V-GPS \cite{nakamoto2024steering}, which reranks actions using a learned value function during retry. As reported in Table~\ref{appendix:tab:baseline}, neither method outperforms \method{} on average. We believe this is because their adaptation remains constrained by the offline data and does not directly learn from the failures encountered at test time, which can be OOD. In contrast, FAR learns directly from test-time failures without requiring human corrections.

In addition, we evaluate DP-FAR-Realtime, which directly compares $s_t$ and $s_{t+H}$ during rollout and detects potential failures based on the change in their estimated values. We find that this approach can cause the robot to repeatedly return to previously visited states, preventing further task progress and limiting the final success rate. Although continuing the rollout may risk failure, it can still allow progress toward task completion. We therefore use hindsight labeling after each attempt for more reliable failure attribution.

\subsection{Additional Ablation Studies}

\begin{table*}[t]
\centering
\small
  \caption{\textbf{Additional Results of Ablation Studies on \method{}.} We evaluate the effectiveness of critic-based positive sample construction, the impact of varying the number of positive samples, as well as the impact of the reward function. The \colorbox{lightgray}{gray-shaded rows} indicate the default settings in each ablation block.
}
  \vspace{-4mm}
  \label{appendix:tab:ablation_FAR}
  \begin{center}
      \setlength{\tabcolsep}{0.8mm}
        \begin{tabular}{l *{7}{S[table-format=2.1(2.1)]}}
          \toprule
          Task  & {PullCube} & {LiftPeg} & {Bread} & {Door}  & \multicolumn{2}{c}{Lift} &  {Average} \\
          \cmidrule(lr){2-2}\cmidrule(lr){3-3}\cmidrule(lr){4-4}\cmidrule(lr){5-5}\cmidrule(lr){6-7}
    $|\mathcal{D}_{\mathrm{exp}}|$ & {\phantom{0}20} & {\phantom{0}20} & {\phantom{0}15} & {\phantom{0}3} & {\phantom{0}5} & {\phantom{0}10} & {\text{--}} \\
          \midrule
          \multicolumn{2}{l}{\textit{Positive Sample Construction}} \\
          \rowcolor{lightgray} Critic Guided & \bf 70.7\pm5.0  & 58.7\pm6.1 & \bf 56.7\pm1.2 & \bf 86.7\pm7.0 & \bf 84.7\pm2.3 & \bf 92.0\pm2.0 & \bf 74.9\pm2.2 \\
          Random & 68.7\pm2.0  & \bf 61.3\pm7.2 & 56.7\pm4.2 & \bf 86.7\pm4.2 & 73.3\pm8.0 & 87.3\pm2.0 & 72.3\pm3.6 \\
          \midrule
          \multicolumn{2}{l}{\textit{Number of Positive Samples}} \\
          $K=1$ & 66.7\pm4.2  & 56.7\pm9.2 & 54.7\pm4.6 & 80.0\pm5.3 & 82.0\pm6.9 & 88.7\pm3.1 & 71.4\pm2.0 \\
          \rowcolor{lightgray} $K=8$ & \bf 70.7\pm5.0  & \bf 58.7\pm6.1 & \bf 56.7\pm1.2 & \bf 86.7\pm7.0 & \bf 84.7\pm2.3 & \bf 92.0\pm2.0 & \bf 74.9\pm2.2 \\
          $K=16$ & 64.7\pm7.6  & 58.0\pm9.2 & 38.7\pm10.1 & 85.3\pm9.0 & 75.3\pm7.6 & 86.0\pm9.2 & 68.0\pm4.7 \\
          \midrule
          \multicolumn{2}{l}{\textit{Reward Type}} \\
          \rowcolor{lightgray} Shaped & \bf 70.7\pm5.0  & 58.7\pm6.1 & \bf 56.7\pm1.2 & \bf 86.7\pm7.0 & \bf 84.7\pm2.3 & \bf 92.0\pm2.0 & \bf 74.9\pm2.2 \\
          Sparse & 68.0\pm4.0  & \bf 61.3\pm12.1 & 55.3\pm4.2 & 83.3\pm9.5 & 83.3\pm7.6 & 85.3\pm2.3 & 72.8\pm2.1 \\
          \bottomrule
        \end{tabular}
  \end{center}
\end{table*}

\textbf{Positive Sample Construction.}
As shown in Table~\ref{appendix:tab:ablation_FAR}, performance drops when distance-based filtering and critic-based selection are removed, and positive samples are instead obtained by directly sampling action chunks from the policy. This indicates that naive sampling does not sufficiently account for the relative quality and diversity of candidate actions.
In particular, actions that are overly similar to the previously failed action should be suppressed, while diverse alternatives with high critic values should be prioritized.

\textbf{Number of Positive Samples.}
As shown in Table~\ref{appendix:tab:ablation_FAR}, the number of positive samples plays an important role in FCPA. Using too few positive samples ($K=1$) provides unstable guidance for model adaptation and therefore limits performance gains. In contrast, using too many positive samples ($K=16$) can also hurt performance, potentially because noisy positives are introduced and the resulting updates contain conflicting optimization directions.

\begin{table*}[ht]
\centering
  \caption{\textbf{Results of Ablation Studies on Retry Perturbation.} We conduct experiments with varying perturbation frequencies and compare the effects of different types of perturbations. The \colorbox{lightgray}{gray-shaded rows} indicate the default settings in each ablation block.}
  \label{appendix:tab:perturb}
  \vspace{-4mm}
  \begin{center}
  \small
      \setlength{\tabcolsep}{1mm}
        \begin{tabular}{l *{7}{S[table-format=2.1(2.1)]}}
          \toprule
          Task  & {PullCube} & {LiftPeg} & {Bread} & {Door}  & \multicolumn{2}{c}{Lift} &  {Average} \\
          \cmidrule(lr){2-2}\cmidrule(lr){3-3}\cmidrule(lr){4-4}\cmidrule(lr){5-5}\cmidrule(lr){6-7}
    $|\mathcal{D}_{\mathrm{exp}}|$ & {\phantom{0}20} & {\phantom{0}20} & {\phantom{0}15} & {\phantom{0}3} & {\phantom{0}5} & {\phantom{0}10} & {\text{--}} \\
          \midrule
          \multicolumn{2}{l}{\textit{Perturbation Frequency}} \\

          $\epsilon_{\text{explore}}=0.0$ & 68.0\pm3.5  & 57.3\pm8.3 & \bf 56.7\pm7.0 & \bf 87.3\pm13.3 & 80.7\pm1.2 & 86.7\pm7.6 & 72.8\pm4.3 \\
          $\epsilon_{\text{explore}}=0.3$ & 68.7\pm4.2  & 56.0\pm12.2 & 54.7\pm8.1 & 83.3\pm10.1 & 77.3\pm8.3 & 90.7\pm1.2 & 71.8\pm3.5 \\
          \rowcolor{lightgray}$\epsilon_{\text{explore}}=0.5$ & \bf 70.7\pm5.0  & 58.7\pm6.1 & \bf 56.7\pm1.2 & 86.7\pm7.0 & \bf 84.7\pm2.3 & 92.0\pm2.0 & \bf 74.9\pm2.2 \\
          $\epsilon_{\text{explore}}=0.7$ & 69.3\pm2.3  & \bf 60.0\pm12.5 & 55.3\pm1.2 & 72.7\pm12.2 & 78.7\pm3.1 & 90.7\pm4.2 & 71.1\pm3.1 \\
          $\epsilon_{\text{explore}}=1.0$ & 70.0\pm6.9  & 58.0\pm6.9 & 55.3\pm1.2 & 72.7\pm8.1 & 80.7\pm7.6 & \bf 94.0\pm4.0 & 71.8\pm1.5 \\
          {Adaptive} & 68.7\pm4.2  & 58.0\pm12.5 & \bf 56.7\pm8.1 & 86.0\pm8.0 & 82.0\pm2.0 & 88.7\pm3.1 & 73.3\pm4.7 \\
          
          \midrule
          \multicolumn{2}{l}{\textit{Perturbation Type}} \\
          \rowcolor{lightgray}{Gaussian} & \bf 70.7\pm5.0  & \bf 58.7\pm6.1 & 56.7\pm1.2 & \bf 86.7\pm7.0 & \bf 84.7\pm2.3 & 92.0\pm2.0 & \bf 74.9\pm2.2 \\
          {Smoothed} & 67.3\pm5.0  & \bf 58.7\pm6.1 & \bf 59.3\pm6.1 & 84.7\pm8.3 & 78.7\pm6.4 & \bf 94.0\pm4.0 & 73.8\pm3.2 \\
          \bottomrule
        \end{tabular}
  \end{center}
\end{table*}

\textbf{Perturbation.}
Table~\ref{appendix:tab:perturb} presents the ablation results for retry perturbation. We first vary the perturbation frequency by adjusting the probability $\epsilon_{\text{explore}}$, while keeping $\Sigma$ fixed across all experiments. Using fewer perturbations improves performance on some long-horizon tasks, such as Door, but reduces performance on shorter-horizon tasks, such as PullCube and Lift. This may be because insufficient perturbation limits exploration and causes the robot to remain stuck in local optima. In contrast, using more perturbations improves performance on shorter-horizon tasks but hurts performance on longer-horizon tasks, possibly because excessive perturbation drives the policy away from the training distribution and limits its effectiveness.

We further evaluate an adaptive perturbation strategy that gradually increases the exploration strength across retries, using
\(\epsilon_{\text{explore}}^{(i)} = 0.3 + 0.1i\),
where \(i\) denotes the retry index starting from 0.
Overall, we find that a moderate fixed perturbation probability,
\(\epsilon_{\text{explore}} = 0.5\), achieves the best trade-off between exploration and exploitation and yields the best overall performance.

We also compare different types of perturbations. Using pure Gaussian noise works relatively well in simulation environments. However, high-frequency noise may damage the robot and cause excessive jitter on a real robot. Compared with pure Gaussian noise, the smoothed perturbation achieves similar performance while being more robot-friendly and safer for real-robot deployment.

\begin{wrapfigure}{r}{0.57\textwidth}
\begin{minipage}{0.55\textwidth}
  \vspace{-0.15in}
  \centering
  \captionsetup[subfloat]{labelformat=empty}
  \subfloat[PullCube $\left | \mathcal{D}_{\mathrm{exp}} \right |=20 $]
  {
\label{appendix:ablation:pullcube}\includegraphics[height=2.7cm]{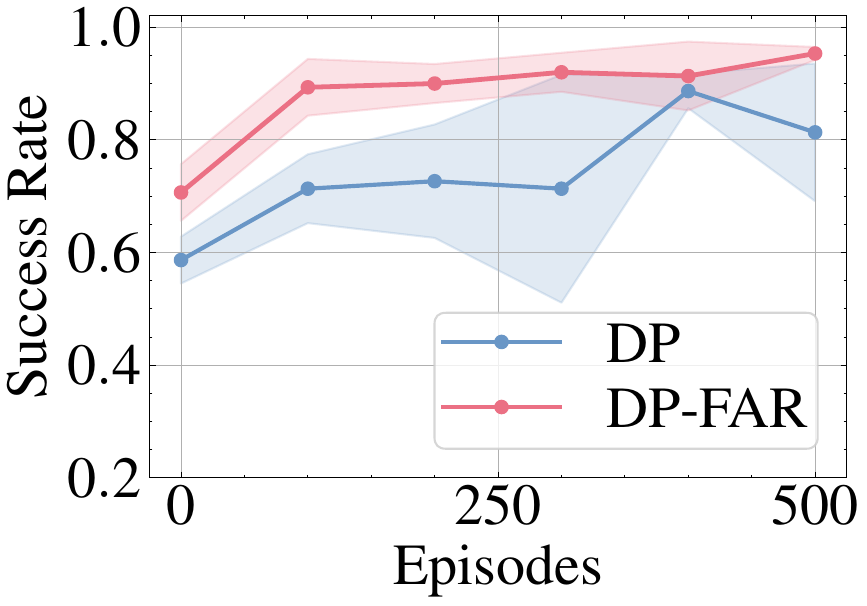} \includegraphics[height=2.7cm]{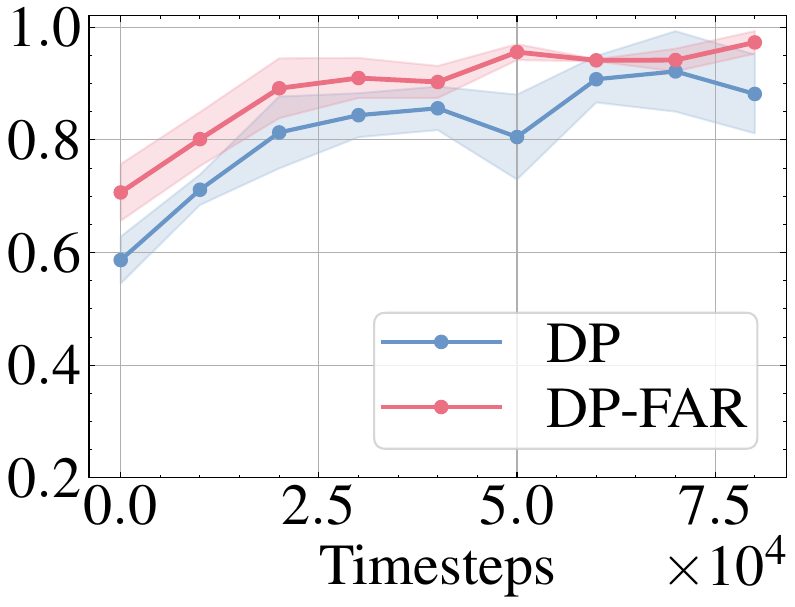}
  }
  \caption{
  \textbf{Continual Policy Improvement under Sparse Reward on PullCube.} Left: success rate versus online episodes. Right: success rate versus environment timesteps.
  With a challenging sparse reward function, \method{} still achieves continual improvement and effectively enhances the data efficiency of online learning.}
  \label{appendix:fig:ablation-sparse}
\end{minipage}
\end{wrapfigure}
\textbf{Reward Type.}
In addition, we examine the impact of the reward function. We replace the shaped reward provided by the environment with a sparse reward, assigning a reward of 3 upon task completion and 0 otherwise. As shown in Table~\ref{appendix:tab:ablation_FAR}, FAR still works well on most tasks with sparse rewards. This is because FAR relies on relative value changes rather than task-specific dense rewards. The results of policy improvement are shown in Figure~\ref{appendix:fig:ablation-sparse}. Even with this extremely sparse reward, \method{} still achieves continual improvement and effectively enhances the data efficiency of online learning. This demonstrates that allocating additional interaction budget to hard failure cases can yield more informative training data, and the newly collected successful trajectories provide a more stable estimate of the optimization direction, ultimately leading to improvements over the standard rollout of the diffusion policy.

\begin{table*}[h]
\centering
  \caption{\textbf{Additional Results of Ablation Studies on FCPA.}
We further conduct ablation studies to examine the design choices of the FCPA algorithm.
Overall, we observe trends consistent with those in the FAR ablations.
The \colorbox{lightgray}{gray-shaded rows} indicate the default settings in each ablation block.}
  \label{appendix:tab:FCPA_ablation}
  \vspace{-4mm}
  \begin{center}
  \small
      \setlength{\tabcolsep}{0.8mm}
        \begin{tabular}{l *{7}{S[table-format=2.1(2.1)]}}
          \toprule
          Task  & {PullCube} & {LiftPeg} & {Bread} & {Door}  & \multicolumn{2}{c}{Lift} &  {Average} \\
          \cmidrule(lr){2-2}\cmidrule(lr){3-3}\cmidrule(lr){4-4}\cmidrule(lr){5-5}\cmidrule(lr){6-7}
    $|\mathcal{D}_{\mathrm{exp}}|$ & {\phantom{0}20} & {\phantom{0}20} & {\phantom{0}15} & {\phantom{0}3} & {\phantom{0}5} & {\phantom{0}10} & {\text{--}} \\
          \midrule
          \multicolumn{4}{l}{\textit{Failure Attribution}} \\
          \rowcolor{lightgray}Value-Percentile & \bf 68.0\pm3.5  & 57.3\pm8.3 & \bf 56.7\pm7.0 & 87.3\pm13.3 & \bf 80.7\pm1.2 & \bf 86.7\pm7.6 & \bf 72.8\pm4.3 \\
          Value-Threshold  & 67.3\pm7.0 & \bf 58.0\pm11.1 & 49.3\pm7.6 & \bf 88.7\pm12.7 & 74.0\pm9.2 & \bf 86.7\pm10.1 & 70.7\pm4.7 \\
          Whole Trajectory  & 65.3\pm6.1  & 52.7\pm7.6 & 54.7\pm9.9 & 85.3\pm15.1 & 76.7\pm9.9 & 83.3\pm6.4 & 69.7\pm5.6 \\
          \midrule
          \multicolumn{4}{l}{\textit{Adaptation Objective}} \\       
          \rowcolor{lightgray}Pairwise Preference  & \bf 68.0\pm3.5  & \bf 57.3\pm8.3 & \bf 56.7\pm7.0 & 87.3\pm13.3 & \bf 80.7\pm1.2 & \bf 86.7\pm7.6 & \bf 72.8\pm4.3 \\
          w/o Positive  & 66.0\pm2.0  & 51.3\pm10.3 & 46.0\pm3.5 & 50.7\pm13.3 & 68.7\pm11.4 & 78.7\pm3.1 & 60.2\pm2.8 \\
          w/o Negative  & 65.3\pm5.0  & 55.3\pm8.3 & 53.3\pm5.0 & \bf 88.0\pm12.2 & 76.0\pm7.2 & 85.3\pm8.3 & 70.6\pm2.9 \\
          AWR  & 64.7\pm1.2  & 56.7\pm9.5 & 51.3\pm5.8 & 84.0\pm14.0 & 64.7\pm14.0 & 83.3\pm5.0 & 67.4\pm3.9 \\
          \midrule
          \multicolumn{2}{l}{\textit{Positive Sample Construction}} \\
          \rowcolor{lightgray} Critic Guided & \bf 68.0\pm3.5  & \bf 57.3\pm8.3 & 56.7\pm7.0 & \bf 87.3\pm13.3 & \bf 80.7\pm1.2 & 86.7\pm7.6 & \bf 72.8\pm4.3 \\
          Random & 64.7\pm4.2  & 54.7\pm9.5 & \bf 58.7\pm4.2 & 80.0\pm17.4 & 67.3\pm10.3 & \bf 88.7\pm6.1 & 69.0\pm3.2 \\
          \midrule
          \multicolumn{2}{l}{\textit{Number of Positive Samples}} \\
          $K=1$ & 65.3\pm4.6  & 54.7\pm10.1 & 54.0\pm8.7 & 82.7\pm14.7 & 74.7\pm7.6 & 86.0\pm8.7 & 69.6\pm2.9 \\
          \rowcolor{lightgray} $K=8$ & \bf 68.0\pm3.5  & \bf 57.3\pm8.3 & \bf 56.7\pm7.0 & \bf 87.3\pm13.3 & \bf 80.7\pm1.2 & \bf 86.7\pm7.6 & \bf 72.8\pm4.3 \\
          $K=16$ & 64.7\pm5.8  & 54.0\pm9.2 & 50.0\pm7.2 & 82.0\pm17.8 & 71.3\pm3.1 & 84.7\pm8.1 & 67.8\pm4.0 \\
          \midrule
          \multicolumn{2}{l}{\textit{Reward Type}} \\
          \rowcolor{lightgray} Shaped & \bf 68.0\pm3.5  & \bf 57.3\pm8.3 & \bf 56.7\pm7.0 & 87.3\pm13.3 & \bf 80.7\pm1.2 & \bf 86.7\pm7.6 & \bf 72.8\pm4.3 \\
          Sparse & 62.7\pm4.2  & 56.0\pm7.2 & 52.7\pm7.0 & \bf 89.3\pm13.6 & 70.0\pm12.0 & \bf 86.7\pm4.6 & 69.6\pm3.0 \\
          \bottomrule
        \end{tabular}
  \end{center}
\end{table*}

\textbf{Ablation on FCPA.}
We further provide ablation results for the FCPA algorithm to enable a more comprehensive evaluation. The results are shown in Table~\ref{appendix:tab:FCPA_ablation}. We observe a similar trend without additional perturbations, suggesting that FCPA does not rely on extra noise-based perturbations.


\subsection{Computational Overhead}

\begin{figure*}[t]
  \begin{center}
    \centerline{\includegraphics[width=0.75\linewidth]{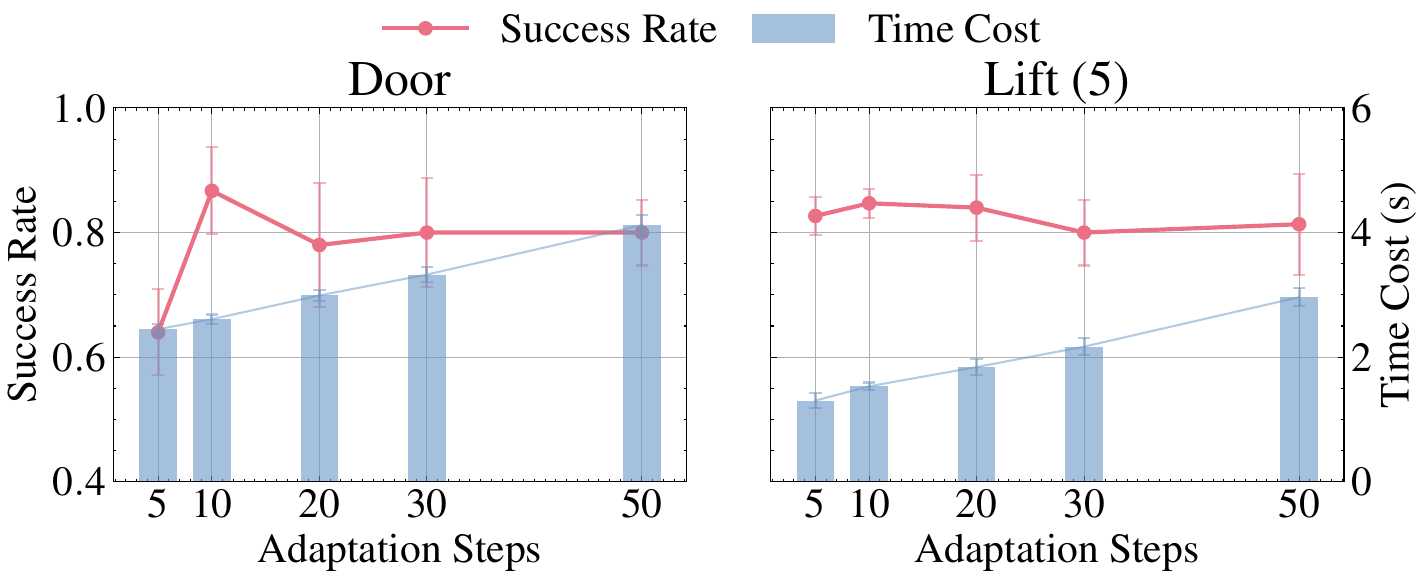}}
      \caption{\textbf{Success Rates and Time Cost vs.\ Adaptation Steps.} Bars show the total adaptation time (s) and lines show the success rate on Door and Lift. A moderate number of adaptation steps performs best, while too many steps degrade performance, likely due to overfitting to the local adaptation data.}
    \label{appendix:fig:ablation-compute}
  \end{center}
\end{figure*}

Figure~\ref{appendix:fig:ablation-compute} reports the total adaptation time for different numbers of TTA steps. Under our default setting of 10 adaptation steps, each adaptation takes approximately 1.5--3 seconds on a single NVIDIA A5000 GPU. We find that a moderate number of adaptation steps yields the best performance, while too many steps degrade performance, potentially due to overfitting to the local adaptation data.



\section{Implementation Details}
\label{appendix:implementation}

\subsection{Network Architecture}
\textbf{Policy Network.}
We adopt a diffusion-based policy as the backbone policy, following the general design of prior diffusion policy methods~\citep{chi2023diffusionpolicy, sailor}. 
At each decision step, the policy is conditioned on a short history of observations, i.e., the previous and current states $[s_{t-1}, s_t]$, and outputs an action chunk $\mathbf{a}_{t:t+H-1}$ with horizon length $H$. 
Each state $s_t$ consists of both proprioceptive measurements and RGB observations.
For visual observations, we employ a ResNet-18 image encoder initialized with \texttt{ImageNet1K\_V1} pretrained weights. 
Following common practice in visuomotor policy learning, the feature maps produced by the encoder are further processed by a Spatial Softmax layer~\citep{robomimic2021} to obtain compact image representations. 
Different camera views, or image inputs at different timesteps, are encoded by separate encoder instances. 
The resulting visual embeddings are then concatenated with the proprioceptive features and used as the observation condition for the denoising network $\epsilon_{\theta}$.
The denoising model is implemented as a conditional 1D U-Net. 
Conditioning information is injected through FiLM layers, similar to the architecture used in Diffusion Policy~\citep{chi2023diffusionpolicy}. 
Specifically, the U-Net is conditioned on both the encoded observation and the diffusion step. 
The diffusion step is first mapped to a 16-dimensional representation using sinusoidal positional encoding, followed by a lightweight MLP. 
The U-Net contains a hierarchy of convolutional and transposed-convolutional blocks with channel dimensions $[64, 128, 256]$ and kernel size 3. 
We use Group Normalization~\citep{wu2018group} and Mish activations~\citep{misra2019mish} throughout the network to improve optimization stability.
Training follows the standard denoising diffusion formulation~\citep{ho2020denoising, nichol2021improved}. 
Given an expert action sequence of length $H$, Gaussian noise is progressively added in the forward diffusion process, and the policy is optimized to predict the injected noise conditioned on the observation and diffusion timestep. 
This corresponds to a behavior cloning objective expressed as a denoising task over action chunks.
At test time, we use DDIM sampling~\citep{song2021denoising} to iteratively denoise a randomly initialized action sequence and obtain the executable action chunk under the current observation.

\textbf{Critic Network.}
The critic module contains two components: a state-value function and an action-value function. 
Both networks share the same observation encoding design as the policy network, while using separate prediction heads on top of the encoded features. 
The value network is equipped with a single linear output head. 
For the action-value network, we employ two independent output heads, following the commonly used double-critic design \citep{fujimoto2018addressing} to mitigate overestimation bias.

\subsection{Training}
Given the offline demonstrations $\mathcal{D}_{\mathrm{exp}}$, we first train a diffusion policy via behavior cloning. We then train the critic models using the same set of demonstrations, initializing the state encoder with the pretrained weights from the diffusion policy. During continual policy improvement, we use the trained diffusion policy to interact with the environment and collect additional training data. We maintain two additional replay buffers, $\mathcal{D}_{\mathrm{succ}}$ and $\mathcal{D}_{\mathrm{fail}}$, which store successful online trajectories (including recovered ones) and failed trajectories, respectively.

In each learning epoch, we collect data from 20 episodes with different initial states. We add Gaussian noise sampled from $\mathcal{N}(0, 0.1^2)$ to the action to encourage exploration while collecting data from the environment. After data collection, we first update the critic models using the aggregated dataset
$\mathcal{D} = \mathcal{D}_{\mathrm{exp}} \cup \mathcal{D}_{\mathrm{succ}} \cup \mathcal{D}_{\mathrm{fail}}$
according to Eq.~(\ref{eq:critic_V}) and Eq.~(\ref{eq:critic_Q}). We then use the updated critic models to compute the advantage of each sample in $\mathcal{D}_{\mathrm{exp}} \cup \mathcal{D}_{\mathrm{succ}}$, and optimize the diffusion policy using Eq.~(\ref{eq:awr}). We use AdamW \citep{loshchilov2018decoupled} for optimization.

\subsection{Evaluation}
During evaluation, we use the same diffusion policy as the base policy and roll it out in the environment.
Within the time limit, if the robot fails to achieve the goal, the attempt is regarded as a failure. In simulation, the environment provides the success label. In real-world experiments, a human observer provides a binary success/failure signal after each attempt.
The robot is then given multiple chances to retry from the failure state. In each retry, only the robot arm is returned to its initial configuration, while the object states and the environment remain unchanged.

Given the previous failure trajectory, \method{} first uses the critic to identify potential action chunks that lead to failure according to Eq.~(\ref{eq:delta_V}), where the percentile is set to 10\% by default.
In Eq.~(\ref{eq:filter}), we set $d_{\min}$ to the average distance between the sampled candidates and the failed action chunk, i.e., $d_{\min} =\bar{d}= \frac{1}{M}\sum_{i=1}^{M} d\!\left(\mathbf a_t^{(i)}, \mathbf a_t^{-}\right)$, so that only candidates deviating more than average from $a_t^-$ are retained.
With the constructed preference pairs, we optimize the denoising network for 10 steps using AdamW, which can be completed within seconds.

\subsection{Hyperparameters}
We list the hyperparameters used in our experiments in Table~\ref{app:tab_hyperparams}. Unless otherwise specified, the same hyperparameters are used across all tasks. Task-dependent hyperparameters are reported as ranges.

\begin{table}
\centering
\caption{\textbf{Hyperparameters used in our experiments.}} \label{app:tab_hyperparams}
\begin{tabular}{l c}
    \toprule
    \textbf{Name} & \textbf{Value} \\
    \midrule
    \textbf{DP Training} & \\
    \midrule
    Batch size &128 \\
    Optimizer & AdamW \\
    Training iterations &30,000 \\
    Online finetuning iterations per epoch & 1,000 \\
    LR scheduler & Cosine annealing \\
    LR scheduler warmup steps & 100 \\
    LR range & [$1\times 10^{-5}$, $1\times 10^{-4}$] \\
    Action chunk horizon $H$ & 8 \\
    Temperature $\eta$ & 1.0 \\
    \midrule
    \textbf{Critic Training} &  \\
    \midrule
    Batch size &128 \\
    Optimizer & AdamW \\
    Training iterations &15,000--30,000 \\
    Online finetuning iterations per epoch & 2,000 \\
    LR & $1\times 10^{-4}$ \\
    Expectile $\tau$ & 0.7 \\
    Discount factor~$\gamma$ & 0.99 \\
    \midrule
    \textbf{Failure-Aware Retry} &  \\
    \midrule
    Percentile $\rho$ & 10\% \\
    Number of action candidates $M$ & 64 \\
    Deviation range $[d_{\min}, d_{\max}]$ & $[\bar{d}, 1.0]$ \\
    Number of positive samples $K$ & 8 \\
    Preference strength $\beta$ & 0.1 \\
    Optimizer & AdamW \\
    LR & $1\times10^{-5}$ \\
    Test-time adaptation steps $n_{\text{tta}}$ & 10 \\
    Perturbation probability $\epsilon_{\mathrm{explore}}$ & 0.5 \\
    Noise std. (per action dimension) & 0.1 \\
    Smoothing coefficient $\alpha$ & 0.3 \\
    \bottomrule
\end{tabular}
\vspace{4pt}
\end{table}

\section{Additional Experimental Details}
\label{appendix:expetiment}

\subsection{Baselines}

\textbf{DP} is the base policy trained via behavior cloning on the offline demonstration dataset, evaluated using standard rollouts.

\textbf{DP-NR} utilizes the same policy model but allows multiple execution attempts without resetting the environment.

\textbf{DP-BGR} is a non-parametric baseline adapted from \citet{du2024err}, which employs rejection sampling to steer actions away from past failures. 
In our implementation, we leverage a trained value function to identify failure-inducing actions by computing the temporal value difference:
\begin{equation}
\Delta V_t = V_{\psi}(s_{t+H}) - V_{\psi}(s_t).
\end{equation}
An action chunk $\mathbf{a}_t$ is classified as failure-inducing if $\Delta V_t$ falls below a predefined threshold. 
The corresponding state-action pair $(s_t, \mathbf{a}_t)$ is then appended to a failure memory. 
During a retry attempt, the agent computes the $L^2$ distance between the current state and all stored failure states in the latent space of the state encoder. 
If the minimum distance falls below a predefined threshold, the recovery mode is triggered; 
the policy then samples 16 candidate action chunks from the diffusion policy and selects the one furthest from the previously failed action. 
This method does not require any parameter updates at test time.

In the policy improvement experiments, we add simple Gaussian action noise to all baselines during data collection to encourage exploration. For a fair comparison, DP-\method{} uses the same action noise during data collection, but additionally applies perturbations during retry to further enhance exploration at test time.

\begin{table}[ht]
	\centering
    \caption{\textbf{Maximum episode lengths (in environment steps) and task descriptions.}
    } \label{app:table_benchmarks}
	\begin{tabular}{lc|cccc}
    \toprule
    Domain & Task & Horizon & Description \\
    \midrule
    \multirow{3}{*}{ManiSkill} 
     & PullCube & 50  & Pull the cube to the target area.\\
     & LiftPeg & 150 & Lift the peg and make it stand upright.\\
     & PokeCube & 100 & Use the tool to poke the cube to the target area.\\
    \midrule
    \multirow{3}{*}{RoboSuite} 
     & Stack & 150 & Lift the red block and place it on top of the other block.\\
     & Door & 200  & Pull down the handle and open the door.\\
     & Bread & 200 & Lift the bread and place it in the correct bin.\\
     \midrule
    \multirow{3}{*}{RoboMimic} 
     & Lift & 100 & Lift the block above the table.\\
     & Can & 200 & Lift the can and place it in the correct bin.\\
     & Square & 200 & Pick up the square tool and insert it into the slot.\\
     \midrule
    \multirow{3}{*}{Real} 
     & Drawer & 200 & Push the handle and close the drawer.\\
     & Pot & 300 & Pick up the lid and place it on the pot.\\
     & Tea & 300 & Pick up the teapot and pour water into the cup.\\
    \bottomrule
\end{tabular}
\end{table}

\subsection{Simulated Experiments}
We consider three manipulation benchmarks: ManiSkill~\citep{taomaniskill3}, RoboSuite~\citep{robosuite2020}, and RoboMimic~\citep{robomimic2021}, covering nine manipulation tasks in total. The task setups are visualized in Figure~\ref{fig:setup}.
Each task provides two RGB observations: one from a third-person camera and one from a wrist-mounted camera. In addition, each task provides proprioceptive states, including the position and orientation of the end effector and the gripper position. The action space is a 7-dimensional vector with values in $[-1, 1]$. The first six dimensions control changes in the position and orientation of the end effector, while the last dimension controls the opening and closing of the gripper. By default, the reward signals used for critic training are the shaped rewards provided by the environments.

For evaluating test-time failure recovery, we run 50 evaluation episodes for each task. The maximum number of attempts per episode is set to 5. We report the average success rate and standard deviation over three random seeds.

For policy improvement experiments, we run 20 episodes to collect data at each training epoch and evaluate the updated policy every 5 epochs. For the Lift, Door, and Can tasks, the maximum number of attempts is set to 5; for the PullCube, LiftPeg, and Square tasks, the maximum number of attempts is set to 3. We report the average success rate and standard deviation over three random seeds.

Table~\ref{app:table_benchmarks} provides the episode horizon and a brief description of each task.

\subsection{Real-world Experiments}
We use the same diffusion policy architecture and evaluate it on three real-world manipulation tasks. All experiments are conducted on a 7-DoF xArm robot with end-effector delta-position control at 30 Hz. The inputs to the policy and critic models include a third-person RGB observation from a ZED Mini camera and the robot proprioceptive state. For each task, we collect 20 successful demonstrations, which are used to train both the policy and critic models. The rewards in the collected datasets are labeled as $-0.1$ before task completion and $3$ upon task completion.

During evaluation, perturbations are applied only to the translational components of the end-effector delta action. We apply the temporal smoothing introduced in Section~\ref{sec:far} to avoid high-frequency jitter. For each task, we evaluate each method over 20 episodes with varying initial configurations to test generalization. The same set of initial configurations is used across different methods for a fair comparison. In each episode, the maximum number of attempts is set to 3 by default. An episode is considered successful if the task is completed within the allowed number of attempts, as determined by human inspection.

\section{Extended Related Work}
\label{appendix:related}

\textbf{Generative Policy Modeling.}
Conventional policy models usually learn a direct mapping from observations to actions with regression losses \citep{pomerleau1988alvinn, zhang2018deep, florence2019self, rahmatizadeh2018vision}. However, such models often struggle to represent the inherently multi-modal action distributions in demonstration data \citep{ibc}.
Prior efforts address this limitation by discretizing the action space and casting policy learning as classification \citep{zeng2021transporter, wu2020spatial}, or by predicting the parameters of simple action distributions such as Gaussians \citep{robomimic2021}. 
Recently, generative policy models, including diffusion-based models \citep{ho2020denoising, song2021denoising} and flow-based and flow-matching models \citep{liu2022flow, lipman2022flow}, have shown strong capability in modeling complex and multi-modal action distributions \citep{chi2023diffusionpolicy}.
These paradigms have also been adopted in large-scale vision-language-action (VLA) models pretrained on diverse robotic datasets \citep{black2024pi_0, black2025pi_05, wen2025diffusionvla, Hou_2025_ICCV_DITA}. By representing a distribution over plausible actions rather than a single deterministic output, such policies are better suited to modeling the diversity and ambiguity inherent in real-world manipulation.

\textbf{Contrastive Learning.}
Contrastive learning has been widely used for learning structured representations \citep{clip, laskin2020curl, qiu2022contrastive}, and has also been applied in reinforcement learning and robotics \citep{clearning, qiu2022contrastive, xu2021positive, ma2024contrastive, lee2025class, fu2025contrastive, eysenbach2022contrastive}. For example, \citet{eysenbach2022contrastive} show that contrastive learning can be cast as goal-conditioned reinforcement learning, with the learned similarity function corresponding to a goal-conditioned value function.
More recently, contrastive objectives have been extended beyond representation learning to preference-based finetuning of generative models. In large language models, Direct Preference Optimization (DPO) \citep{dpo} aligns pretrained models using paired preference data. Similar ideas have also been explored in diffusion models, where Diffusion-DPO \citep{wallace2024diffusion} finetunes pretrained diffusion models from human-labeled preference pairs.
These works suggest that supervision from both preferred and dispreferred samples provides an effective way to reshape a pretrained model's output distribution. This is particularly relevant in our setting, where adaptation must be performed from limited test-time experience: using only negative feedback from previous failures can easily destabilize the policy, while contrastive supervision provides a more balanced update signal. Inspired by contrastive and preference learning, our method uses failed actions as negative samples and alternative actions as positive samples to perform failure-aware adaptation at test time.

\section{Qualitative Results}
\label{appendix:qualitative}
\begin{figure*}[t]
  \begin{center}
    \centerline{\includegraphics[width=0.87\linewidth]{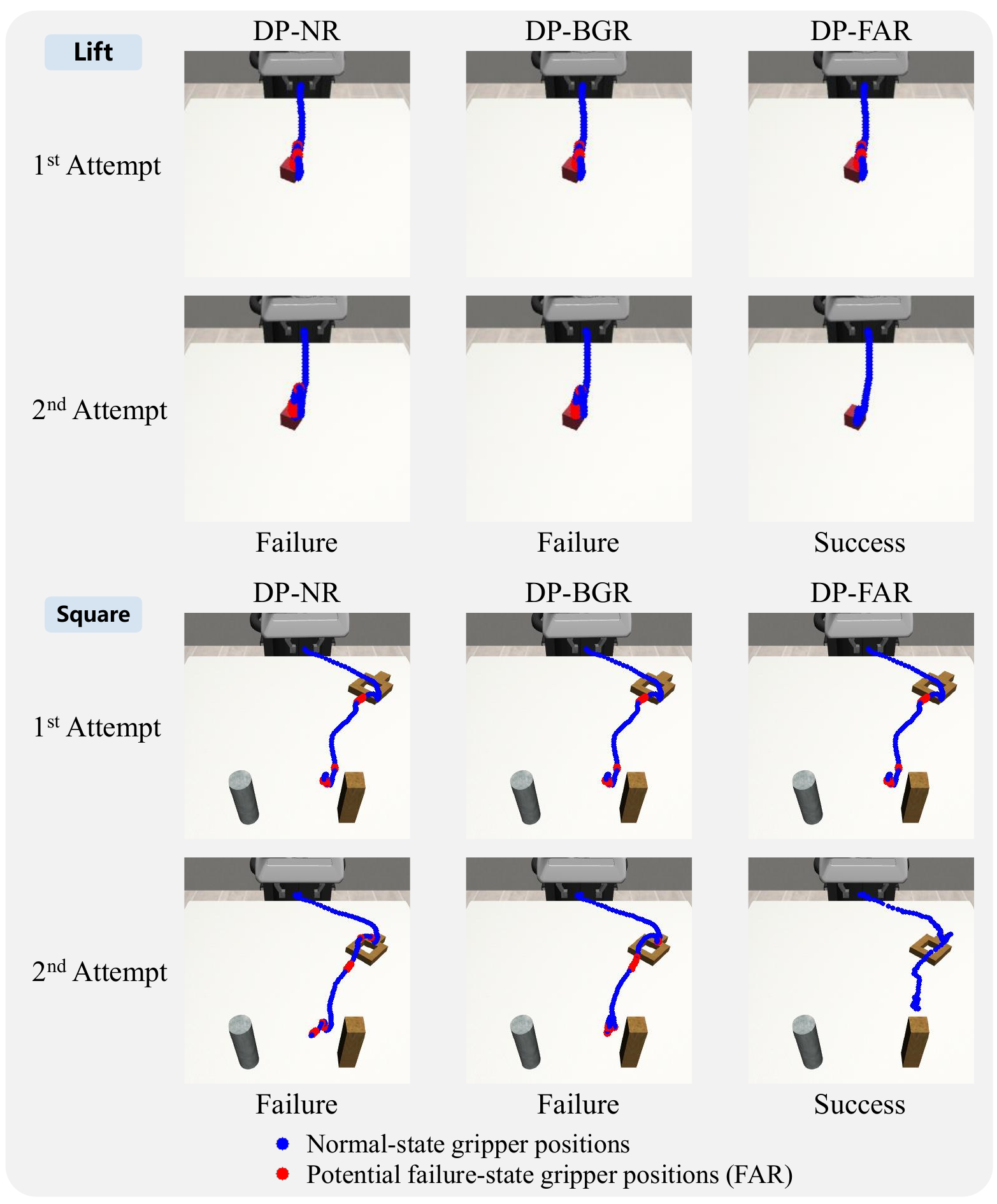}}
    \caption{
    \textbf{Qualitative Results on Lift and Square.} We visualize the gripper-center trajectory, where red and blue points indicate potentially failure-inducing and normal positions identified by FAR.
    }
    \label{appendix:fig:qualitative}
  \end{center}
\end{figure*}

We present additional qualitative results in Figure~\ref{appendix:fig:qualitative}.

